\documentclass{article}

\usepackage{microtype}
\usepackage{graphicx}
\usepackage{caption}

\usepackage{hyperref}

\usepackage[accepted]{icml2024}

\usepackage{amssymb}

\usepackage[textsize=tiny]{todonotes}

\usepackage{url}
\usepackage{subcaption}

\usepackage[ruled, vlined, nofillcomment, linesnumbered, algo2e]{algorithm2e}
\usepackage{pgfplotstable, pgfmath} 
\usepackage{pgfplots} 
\usepackage{enumitem}
\usepackage{siunitx}
\usepackage{thmtools}  
\usepackage{multirow}
\usepackage{makecell}

\usepackage[noenumitem,notheorems]{eda} 
\usepackage{prettyplots} 
\SetKw{Continue}{continue}
\SetKw{Break}{break}
\graphicspath{ {./figs/} }
\pgfplotsset{compat=1.17} 
\usetikzlibrary{calc} 
\usetikzlibrary{backgrounds}
\usepgfplotslibrary{colorbrewer} 
\usetikzlibrary{pgfplots.colormaps} 

\pgfdeclarelayer{background layer}
\pgfdeclarelayer{foreground layer}
\pgfdeclarelayer{midway layer}
\pgfsetlayers{background layer,midway layer,main,foreground layer}

\newif\iffairness
\fairnessfalse
\newif\ifnotes
\notestrue

\newcommand\thefontsize[1]{{#1 The current font size is: \f@size pt\par}}
\usepackage[notheorems]{jpk}

\usepackage{dsfont} 
\usepackage{relsize} 
\usepackage{mathtools} 

\newlist{compactenum}{enumerate}{4}
\setlist[compactenum,1]{nolistsep,nosep,label=(\roman*)}
\newlist{tidyenum}{enumerate}{2}
\setlist[tidyenum]{wide,labelindent=0pt,labelwidth=!,itemsep=.6ex,label=\arabic*.}

\ifnotes
\usepackage{pdfcomment}
\newcounter{jtodoboxes}
\newcommand{\atodo}[2][]{%
  \expandafter\let\expandafter\todobox%
  \csname jtodobox\roman{jtodoboxes}\endcsname%
  \newsavebox\todobox%
  \sbox\todobox{\parbox{5cm}{\large #2}}%
  \edef\todoboxexp{\usebox\todobox}%
  \pdffreetextcomment[%
  width=\wd\todobox,
  height=2\ht\todobox,
  opacity=.6,type=callout,color=white,fontcolor=red,#1%
  ]{#2}%
}
\colorlet{annotations janis}{blue!20!white}
\newcommand{\anote}[2][]{%
  \rlap{\pdfcomment[%
  author=Janis,icon=Comment,opacity=.5,color=annotations janis,#1%
  ]{#2}}%
}
\newcommand{\asent}[1][sentinel]{\textrule[colour=annotations janis,tag text=Janis,tag colour=blue]{#1}}
\newcommand*{\aanot}[1]{\textcolor{blue}{#1}}

\newcommand*{\atag}[1]{\ttag[blue]{#1}}
\makeatletter
\def\uwave{\bgroup \markoverwith{\lower3.5\p@\hbox{\sixly \textcolor{red}{\char58}}}\ULon}
\font\sixly=lasy6 
\makeatother

\else

\newcommand*{\aanot}[1]{}
\newcommand{\atodo}[2][]{}
\newcommand{\atag}[1]{}
\newcommand{\anote}[2][]{}
\newcommand{\asent}[1][]{}

\renewcommand{\note}[1]{}
\fi 

\icmltitlerunning{Learning Exceptional Subgroups by End-to-End Maximizing KL-divergence}
\begin{document}




\newcommand{\lreals}{\mathbb{R}} 
\newcommand{\lfeatel}{x} 
\newcommand{\lfeat}{\fvec\lfeatel} 
\newcommand{\lrfeat}{\fvec X} 

\newcommand{\ltarg}{y} 
\newcommand{\ltargv}{\fvec\ltarg} 
\newcommand{\lrtarg}{Y} 
\newcommand{\lrtargv}{\fvec Y} 
\newcommand{\lrsel}{S} 
\newcommand{\ltargspace}{\mathcal{Y}} 
\newcommand{\lfeatspace}{\lreals^\lnfeat} 
\newcommand{\lfeatspacep}{\lfeatspace_{\in}} 
\newcommand{\lfeatspacem}{\lfeatspace_{\notin}} 
\newcommand{\lfeatspacems}{\lfeatspace_{\notin\subset}} 

\newcommand{\lrule}{\sigma} 
\newcommand{\lpred}{\pi} 
\newcommand{\xpred}[3]{\lpred(#1;#2,#3)} 

\newcommand{\llb}{\alpha} 
\newcommand{\lub}{\beta} 
\newcommand{\flb}[1]{\llb_{#1}} 
\newcommand{\fub}[1]{\lub_{#1}} 
\newcommand{\findic}[1]{\mathds{1}\!\left\{#1\right\}} 
\newcommand{\xkldiv}[2]{D_{\mathrm{KL}}\left(#1\middle\|#2\right)}
\newcommand{\wkldiv}[2]{D_{\mathrm{WKL}}\left(#1\middle\|#2\right)}

\newcommand{\lpdftarg}{\fpdf\lrtarg} 
\newcommand{\lpdftargsg}{\fpdf{\lrtarg|\lrsel=1}} 
\newcommand{\lcdftarg}{\fcdf\lrtarg} 
\newcommand{\lcdftargsg}{\fcdf{\lrtarg|\lrsel=1}} 
\newcommand{\lpdfsel}{\fpdf{\lrsel=1|\lrfeat}} 
\newcommand{\lnfeat}{m} 
\newcommand{\lnargs}{t} 
\newcommand{\lnsamp}{n} 
\newcommand{\lnpsg}{k} 

\newcommand{\lprb}{\mathbb{P}} 
\newcommand{\lpdf}{p} 
\newcommand{\lcdf}{P} 

\newcommand{\lweisize}{\gamma} 

\newcommand{\lprule}{s} 
\newcommand{\indep}{\perp \!\!\! \perp}

\newcommand{\ldef}{\coloneqq} 

\newcommand{\lpartemp}{t} 

\newcommand{\fvec}[1]{\mathbf{#1}} 
\newcommand{\fprb}[1]{\lprb\left(#1\right)} 
\newcommand{\fpdf}[1]{\lpdf_{\mathsmaller{#1}}} 
\newcommand{\fcdf}[1]{\lcdf_{\mathsmaller{#1}}} 
\newcommand{\fdef}[1]{\textbf{#1}} 
\newcommand{\fdesc}[1]{``\emph{#1}''} 
\newcommand{\xexp}[2][]{\mathop{{}\mathlarger{\mathbb{E}}}\ifthenelse{\equal{#1}{}}{}{_{#1}}\left[#2\right]} 

\newcommand{\falg}[1]{\texttt{#1}\xspace} 
\newcommand{\lmbh}{\falg{BumpHunting}} 
\newcommand{\lmwrac}{\falg{SD-}$\mu$} 
\newcommand{\lmmdl}{\falg{SD-MDL}} 

\newcommand{\ftag}[1]{{\small[\textsc{#1}]}} 


\newcommand{\jc}[1]{{\color{orange} (JC: #1)}}

\newcommand{\ourmethod}{$\textsc{Syflow}$\xspace}

\newcommand{\bh}{$\textsc{Bh}$\xspace}

\newcommand{\rsd}{$\textsc{Rsd}$\xspace}

\newcommand{\sd}{$\textsc{Sd}\text{-}\mu$\xspace}
\newcommand{\sdkl}{$\textsc{Sd}\text{-KL}$\xspace}
\newcommand{\ssd}{\sd}
\newcommand{\mywedge}{\; \wedge \;}

\newtheorem{theorem}{Theorem}
\newtheorem{conjecture}{Conjecture}
\newtheorem{lemma}{Lemma}
\newtheorem{corollary}{Corollary}
\newtheorem{postulate}{Postulate}
\newtheorem{proposition}{Proposition}
\newtheorem{assumption}{Assumption}
\newtheorem{definition}{Definition}
\newtheorem{example}{Example}
\newenvironment{proof}{\paragraph{Proof:}\itshape}{\hfill$\square$ \\}

\def\Var{{\mathrm{Var}}}
\def\Cov{{\mathrm{Cov}}}
\def\E{{\mathrm{E}}\,}
\def\Effect{{\text{Effect}}}
\def\V{{\text{V}}}
\def\R{{\mathbb{R}}}
\def\bin{{\text{softbin}}}
\def\pred{{\hat{\pi}}}
\let\learner\lprule
\def\flow{{\text{flow}}}
\newcommand{\mean}{\ensuremath{n_s}\xspace}
\newcommand{\BC}{\ensuremath{\mathit{BC}}\xspace}
\newcommand{\DKL}{\ensuremath{D_{\mathit{KL}}}\xspace}
\newcommand{\AMD}{\ensuremath{\mathit{AMD}}\xspace}

\newcommand{\Indep}{\mathop{\perp\!\!\!\perp}\nolimits} 
\newcommand{\nIndep}{\mathop{\cancel\Indep}\nolimits}

\def\ourscore{\mathit{Score}}

\makeatletter
\renewcommand*{\@fnsymbol}[1]{\ensuremath{\ifcase#1\or   \circ\or \bullet\or *\or \ddagger\or
		\mathsection\or \mathparagraph\or \|\or **\or \dagger\dagger
		\or \ddagger\ddagger \else\@ctrerr\fi}}
\makeatother

\newcounter{inlineequation}
\setcounter{inlineequation}{0}
\renewcommand{\theinlineequation}{(\Roman{inlineequation})}

\newcommand{\inlineeq}[1]{\refstepcounter{inlineequation}\theinlineequation\ \(#1\)}

\definecolor{ind-effect}{HTML}{1B77B8}
\definecolor{pos-effect}{HTML}{25A012}
\definecolor{neg-effect}{HTML}{E05263}

\pgfplotscreateplotcyclelist{ishap-stacked-ybar}{%
	{ind-effect, fill=ind-effect},
	{pos-effect, fill=pos-effect},	
	{neg-effect, fill=neg-effect},
}

\definecolor{n1}{HTML}{1B77B8}
\definecolor{n2}{HTML}{FF7E00}
\definecolor{n3}{HTML}{25A012}
\definecolor{n4}{HTML}{D82520}
\definecolor{n5}{HTML}{9567C1}
\definecolor{n6}{HTML}{8D5649}
\definecolor{n7}{HTML}{E476C5}
\definecolor{n8}{HTML}{7F7F7F}
\definecolor{n9}{HTML}{BCBD00}
\definecolor{n10}{HTML}{00BED1}

\pgfplotscreateplotcyclelist{nshap-double}{%
	{n1, fill=n1},
	{n1, fill=n1},
	{n2, fill=n2},
	{n2, fill=n2},
	{n3, fill=n3},
	{n3, fill=n3},
	{n4, fill=n4},
	{n4, fill=n4},
	{n5, fill=n5},
	{n5, fill=n5},
	{n6, fill=n6},
	{n6, fill=n6},
	{n7, fill=n7},
	{n7, fill=n7},
	{n8, fill=n8},
	{n8, fill=n8},
	{n9, fill=n9},
	{n9, fill=n9},
	{n10, fill=n10},
	{n10, fill=n10},
}

\pgfplotsset{
	my legend/.style={
		mark=square*,
		fill,
		mark options={scale=4, fill opacity=0.5}
	}
}

\newcommand{\scalabilityLinePlot}[1]{
\begin{subfigure}[b]{0.33\textwidth}
	\begin{tikzpicture}
		\usetikzlibrary{calc}
		\begin{axis}[ 
			pretty line,
			pretty xlabeltilt,
			cycle list name = prcl-line,
			legend style={xshift=4.35cm,yshift=0.65cm},
			pretty labelshift,
			width = \linewidth,
			height=3cm,
			ylabel={F1-score},
			xlabel={Number of features $|m|$},
			ymax=1.0,
			legend columns = 7,
			tick label style={/pgf/number format/fixed}
		] 
		\foreach \x in {flows, bh,  rsd, sd_kl, ssd}{
			\addplot table[x={Dim},y={F1 Score},col sep=comma,y error=Std]{expres/scalability/\x_scalability_#1.csv};
		}      
		\end{axis}
	  \end{tikzpicture}
	  \subcaption{F1-score for $f_p=#1$}
	  \label{fig:exp1b}
	\end{subfigure}%
}
\newcommand{\scalabilityLinePlotTest}[1]{
\begin{subfigure}[b]{0.33\textwidth}
	\begin{tikzpicture}
		\usetikzlibrary{calc}
		\begin{axis}[ 
			pretty line,
			pretty xlabeltilt,
			cycle list name = prcl-line,
			legend style={xshift=4.35cm,yshift=0.65cm},
			pretty labelshift,
			width = \linewidth,
			height=3cm,
			ylabel={F1-score},
			xlabel={Number of features $|m|$},
			ymax=1.0,
			legend columns = 7,
			tick label style={/pgf/number format/fixed}
		] 
		\addplot+[error bars/.cd, y dir = both, y explicit] table[x={Dim},y={F1 Score},col sep=comma,y error=Std]{expres/scalability/flows_scalability_#1_test.csv};
		\foreach \x in { bh,  rsd, sd_kl, ssd}{
			\addplot+[error bars/.cd, y dir = both, y explicit] table[x={Dim},y={F1 Score},col sep=comma,y error=Std]{expres/scalability/\x_scalability_#1.csv};
		}      
		\end{axis}
	  \end{tikzpicture}
	  \subcaption{F1-score for $f_p=#1$}
	  \label{fig:exp1b}
	\end{subfigure}%
}
\newcommand{\scalabilityLinePlotWithLegend}[1]{
\begin{subfigure}[b]{0.33\textwidth}
	\begin{tikzpicture}
		\usetikzlibrary{calc}
		\begin{axis}[ 
			pretty line,
			pretty xlabeltilt,
			cycle list name = prcl-line,
			legend style={xshift=7cm,yshift=0.9cm},
			width = \linewidth,
			height=3cm,
			ylabel={F1-score},
			xlabel={Number of features $|m|$},
			ymax=1.0,
			legend columns = 7,
			legend entries = {\ourmethod,\bh ,\rsd, \sdkl,\ssd},
			tick label style={/pgf/number format/fixed}
		] 
		\foreach \x in {flows, bh,  rsd, sd_kl, ssd}{
			\addplot table[x={Dim},y={F1 Score},col sep=comma]{expres/scalability/\x_scalability_#1.csv};
		}      
		\end{axis}
	  \end{tikzpicture}
	  \subcaption{F1-score for $f_p=#1$}
	  \label{fig:exp1b}
	\end{subfigure}%
}

\twocolumn[


\icmltitle{Learning Exceptional Subgroups by End-to-End Maximizing KL-divergence}
\icmlsetsymbol{equal}{*}

\begin{icmlauthorlist}
\icmlauthor{Sascha Xu}{equal,cispa}
\icmlauthor{Nils Philipp Walter}{equal,cispa}
\icmlauthor{Janis Kalofolias}{cispa}
\icmlauthor{Jilles Vreeken}{cispa}
\end{icmlauthorlist}

\icmlaffiliation{cispa}{CISPA Helmholtz Center for Information Security}

\icmlcorrespondingauthor{Sascha Xu}{sascha.xu@cispa.de}
\icmlcorrespondingauthor{Nils Philipp Walter}{nils.walter@cispa.de}

\icmlkeywords{Subgroup Discovery, Normalizing Flows}

\vskip 0.3in
]
\printAffiliationsAndNotice{\icmlEqualContribution} 

\begin{abstract}
  Finding and describing sub-populations that are exceptional regarding a target property
has important applications in many scientific disciplines, from identifying
disadvantaged demographic groups in census data to finding conductive molecules within gold nanoparticles.
Current approaches to finding such \emph{subgroups} require pre-discretized predictive variables, 
do not permit non-trivial target distributions, do not scale to large datasets, and struggle to find diverse results. 

To address these limitations, we propose \ourmethod, an end-to-end optimizable approach in which we 
leverage normalizing flows to model arbitrary target distributions, and introduce a novel neural layer that results in
easily interpretable subgroup descriptions.
We demonstrate on synthetic and real-world data, including a case study,
that \ourmethod reliably finds highly exceptional subgroups accompanied by insightful descriptions.

\end{abstract}

\section{Introduction}
\input{texfigs/gold_subgroup}
The majority of modern machine learning focuses on finding global models that
perform well on \emph{predictive} tasks such as classification.
Here, deep neural networks often achieve state-of-the-art performance, at the expense of human-interpretable insight.

Orthogonal to the advances in predictive modeling, many scientific applications require \emph{descriptive} modeling;
finding sub-populations that are somehow \emph{exceptional}, and providing a human-interpretable description for these. 
Applications of finding such \emph{subgroups} range from identifying disadvantaged demographic groups in census data ~\citep{boll:19:paygap,ortiz:11:inequality} 
to learning those combinations of properties that single out materials with desirable properties \citep{sutton:20:subgroup-material}.

The common denominator in such applications is to present the relevant subgroups to a domain expert. 
In other words, not only do we require to find subsets with exceptional behavior, but also, that these can clearly be interpreted by the respective audience. 
That is, we have a joint optimization task of learning simple descriptions of sub-populations for which the property of interest is (locally) exceptionally distributed compared to the rest of the dataset. Typically, such a description is a conjunction of predicates, each based on the features of 
the dataset. For example, on census data with \emph{wage} as the target property, a subgroup description 
could be \fdesc{Women without higher education} (Fig.~\ref{fig:wage-selector}), where we find that their salary is exceptionally low (Fig.~\ref{fig:wage-distribution}).

Since the introduction of subgroup discovery by~\citeauthor{klosgen:96:explora}, several approaches have been proposed~\citep{atzmueller:15:subgroup}.
However, they have not kept up with the recent advances in machine learning. 
Prior approaches suffer from three main limitations. 
First, due to combinatorial optimization, these methods are limited to small datasets.
Second, most methods assume that the target follows a simple, e.g. normal or binomial distribution. 
Although there are proposals to learn a proxy of the target distribution, their results are less interpretable.
Third, previous methods  require a pre-quantisation of the continuous features, which is independent of the optimization procedure.
As we show in our experiments, this greatly influences the quality of the results.

To overcome these limitations, we propose \ourmethod, where we introduce three major changes:
\begin{compactenum}
    \item We formulate subgroup discovery as a continuous optimization problem based on KL-divergence. 
This enables first-order optimization, which significantly improves runtime and performance.
    \item We leverage Normalizing Flows~\citep{rezende:15:flows}
		  to accurately learn the target distribution from data, enabling us to deal with intricate real-world distributions.
    \item To learn interpretable subgroup descriptions, we propose a neuro-symbolic rule learner that in an end-to-end fashion learns the conjunction predicates
that best describe a subgroup as well as the corresponding discretization.
\end{compactenum}

We extensively evaluate \ourmethod on synthetic and real-world data, comparing against state-of-the art methods.
We show that \ourmethod accurately and reliably learns and characterizes exceptional subgroups, even for complex target distributions.
We demonstrate that in contrast to the baselines, \ourmethod identifies a diverse set of subgroups.
To showcase  \ourmethod's strength on real-world tasks, we perform a case study in the domain of materials science, where
we search for characteristics  of gold clusters with certain conductivity (Fig.~\ref{fig:homo-lumo}) and reactivity.
In both, \ourmethod identifies physically plausible subgroups and their descriptions.
%

\section{Preliminaries}
We consider a dataset of $\lnsamp$ pairs $(\lfeat,\ltarg)$, where $\ltarg\in\ltargspace$ represents a property of interest, the \fdef{target property}, and $\lfeat\in\lfeatspace$ is a \fdef{feature vector}. From a statistical perspective, we assume $(\lfeat,\ltarg)$ is a realisation of a pair of random variables $(\lrfeat,\lrtarg)\sim\fprb{\lrfeat,\lrtarg}$. 
We denote random variables by capitals, write $\lpdf$ for their density, and $\lcdf$ for their law. 

We are interested in learning \textbf{subgroups} for which the conditional distribution of the target attribute $P_{Y|S=1}$ is exceptional compared to $\lcdftarg$. 
A subgroup membership function, or \fdef{rule}, $\lrule(\lfeat) \in \{0,1\}$ determines whether a sample $\lfeatel$ belongs to the subgroup (1) or not (0). Formally, it is a conjunction $\lrule:\lfeat\mapsto \land_{i=1}^\lnfeat \pi(\lfeatel_i)$ of Boolean-valued predicates $\lpred:\lreals\to\{0,1\}$, where each predicate defines an interval over which its output is 1, e.g. ``$18 > \mathrm{age} > 65$''. 

To permit continuous optimization, we consider \textbf{soft predicates} $\hat\pi \in [0,1]$. These are smooth functions that model the probability of a sample $x$ to be inside an interval $\alpha < x < \beta$. We can control the steepness of the transitation via a temperature parameter $t$. We write $s(x) \in [0,1]$ to denote a \textbf{soft rule} based on soft predicates, and $S \in \{0,1\}$ for the indicator (random) variable that $s$ defines, i.e. $\lprb(\lrsel=1|\lrfeat=\lfeatel)=\lprule(\lfeatel)$. 
Note that by reducing $t$ to $0$ we again obtain binary predicates and rules that are easy to interpret.


\section{Method}
\label{sec:theory}

In this section we introduce \ourmethod for learning exceptional subgroups by end-to-end maximization of KL-divergence. We first give an overview and then the details.

\subsection{Overview}

A subgroup is characterized by a membership function $\lrule$, which is commonly constrained to be a directly interpretable rule $\lrule$, i.e.~a logical conjunction over predicates $\lpred$. Finding the rule that identifies the most exceptional conditional distribution of the target is an NP-hard combinatorial problem~\cite{boley_non-redundant_2009}.

We propose to take a different, end-to-end optimizable approach to learning subgroups. To this end we propose a continuous relaxation $\lprule$ of the binary-valued rule function, which is designed to be differentiable, avoiding the need for pre-discretization, yet directly give interpretable results. We propose to optimize the exceptionality of a subgroup in terms of KL-divergence between the conditional distribution $P_{Y|S=1}$ and the marginal distribution $P_Y$ of the target, where we model these distributions non-parametrically using Normalizing Flows -- with the added benefit that we have a single solution for univariate or multivariate targets.

As our entire framework is differentiable, we can optimize all components with gradient descent, which as we will see is often both faster and more performant than combinatorial approaches. Finally, our framework naturally enables iteratively learning multiple non-redundant subgroups by regularizing with the similarity of already learned subgroups.

\subsection{Differentiable Rule Induction}
\label{sec:rule-induction}
\input{texfigs/boxfig.tex}
Subgroup membership is defined by a logical conjunction over binary predicates $\lrule:\lfeat\mapsto \land_{i=1}^\lnfeat \xpred {\lfeatel_i} {\flb i} {\fub i}$.
Predicates on features that are not relevant to a rule are set to be always true.

Our key idea is to redefine a subgroup membership function in probabilistic terms, such that we obtain a function that acts akin to logical conjunctions but at the same time are differentiable and therewith optimizable using gradient descent.  
In particular, we propose to associate each feature $X_i$ with a \fdef{soft predicate}
\begin{align}
    \pred(\lfeatel_i;\llb_i,\lub_i,\lpartemp) = \frac{e^{\frac{1}{t} (2\lfeatel_i - \llb_i)}}{e^{\frac{1}{t}\lfeatel_i} + e^{\frac{1}{t}(2 \lfeatel_i - \llb_i)} + e^{\frac{1}{t} (3 x_i - \llb_i - \lub_i)}} \; ,
    \label{eq:predicate}
\end{align}
where we adapt the idea of approximate and differentiable splits from deep decision trees~\cite{yang:18:deeptree} to use a lower and upper bound $\llb, \lub\in\lreals$, and introduce a temperature parameter $\lpartemp>0$ that controls the steepness of the function at these bounds. The lower the temperature, the less soft the predicate, and in the limit of $\lpartemp \to 0$ a soft predicate converges to a strict predicate. In Fig.~\ref{fig:soft-binning} we show an example soft predicate for different temperatures.

\begin{restatable}{theorem}{theorembin}
    \label{eq:binning}
    Given its lower and upper bounds $\llb_i, \lub_i \in \R$, the soft predicate of Eq.~\eqref{eq:predicate} applied on $\lfeatel \in R$ converges to the crisp predicate that decides whether $\lfeatel\in(\llb,\lub)$,
    \[
        \lim_{\lpartemp \to 0} \pred(x_i;\llb_i,\lub_i,\lpartemp) = \begin{cases}
            1 & \text{if } \llb_i < x_i < \lub_i  \\
            0.5 & \text{if}\, x_i = \llb_i \lor x_i = \lub_i \\
            0 & \text{otherwise}
        \end{cases}\;.
    \]
\end{restatable}
We provide the full proof for the general case with $M$ bins in the Appendix \ref{ap:binproof}.

The soft predicate $\pred$ provides a differentiable, adaptable binning function. 
Next, we propose to combine the predicates $\pred_i$ for each feature $x_i$ into a soft rule $\lprule$ that acts akin to logical conjunctions but remains differentiable. It is possible to model a logical conjunction using multiplication, but this leads to vanishing gradients \cite{hochreiter:98:vanishing}, which is problematic especially for non-trivial amounts of soft predicates (features). The harmonic mean 
\begin{equation}
    \label{eq:harmonic-mean}
\mathcal{M}(x) = \frac{p}{\sum_{i=1}^\lnfeat \pred(x_i;\llb_i,\lub_i,\lpartemp)^{-1}} \; ,
\end{equation}
behaves as desired for strictly binary predicates, i.e.~$\exists \pred(x_i;\llb_i,\lub_i,\lpartemp)=0\Rightarrow \mathcal{M}(x) = 0$ and $\forall \pred(x_i;\llb_i,\lub_i,\lpartemp)=1\Rightarrow \mathcal{M}(x) = 1$, but tends to break down when given many soft predicates, e.g.~when considering a high dimensional feature space. 
To avoid this, we propose to instead use the \textit{weighted harmonic mean} to model logical conjunctions, and construct the soft rule function $\lprule$ as
\begin{equation}
    \label{eq:soft-predicate}
    \learner(x;\llb,\lub,a,\lpartemp) = \frac{\sum_{i=1}^\lnfeat a_i}{\sum_{i=1}^\lnfeat a_i \pred(x_i;\llb_i,\lub_i,\lpartemp)^{-1}}\;.
\end{equation}
The weights $a \in \R^\lnfeat$, which are constrained to be positive through a ReLU function, allow the conjunction layer to disable unnecessary predicates.
The weights do not affect behavior for strictly binary predicates, wherever $a_i>0$.

We show an example subgroup membership function for a soft rule $\lprule$ in Fig.~\ref{fig:membership-box}. The subgroup here is characterized by a conjunction of two predicates on $X_1$ and $X_2$, i.e.~a box, with a gradual, smooth transition from subgroup inclusion to 
exclusion at the boundaries.

In general, our formulation of a soft rule is completely flexible in regards to the thresholds of the binned features, and asymptotically for $\lpartemp \to 0$ is equivalent to a strict rule. 

\subsection{Differentiable Density Estimation}
\label{sec:flows}
Besides a differential rule function, we require accurate estimations of the conditional resp. marginal distributions of the target. We deviate from existing work by taking a differentiable non-parametric approach in the form of \textit{Normalising Flows}. These are an increasingly popular class of density estimators \citep{rezende:15:variational,dinh:17:density,papamakarios:21:normalizing}. The fundamental idea behind a normalising flow is to start with a distribution with a known density function, e.g. a Gaussian distribution with $\fpdf{\mathcal{N}}$,
and fit an invertible function $f$ to transform it onto the target density.

Our method, \ourmethod, allows to seamlessly use any normalising flow architecture. In this work, our architecture of choice are Neural Spline Flows \citep{durkan:19:splineflows}, which use expressive yet invertible piece-wise, polynomial spline functions.
In general, the idea is to train the function $f$ so that $\fpdf{\ltargspace}\approx f(\fpdf{\mathcal{N}})$.
Given a sample $\ltarg$, we can compute the likelihood of that point under the current function $f$ as 
$\fpdf{f(\mathcal{N})}(\ltarg)=\fpdf{\mathcal{N}}(f^{-1}(\ltarg))|\det\left(\frac{\delta f^{-1}(\ltarg)}{\delta \ltarg}\right)|$.
Thus, given a sample of $\fprb{\lrtarg}$, we can maximise the likelihood of $\fpdf{f(\mathcal{N})}(\ltarg)$ and hence fit
$\fpdf{f(\mathcal{N})} \approx \fpdf{\ltargspace}$.


\subsection{Differentiable Exceptionality Measure}
\label{sec:outstanding}
We now have a differentiable rule function $\lprule$ and versatile density estimate $\lpdftarg$ resp. $p_{Y|S=1}$ of the target distribution.
Next, we propose a differentiable measure of exceptionality between the conditional target distribution and the marginal target distribution. We adopt the Kullback-Leibler (KL) divergence and show how we can reformulate it towards this goal.
We start with the standard definition, 
\begin{equation}
    \label{eq:KL} 
    \xkldiv \lcdftargsg \lcdftarg = \hspace{-2pt}\int_{\ltarg \in \ltargspace} \hspace{-2pt}\lpdftargsg(\ltarg) \log\left(\frac{\lpdftargsg(\ltarg)}{\lpdftarg(\ltarg)}\right) d\ltarg\;.
\end{equation}
Here, the soft rule $\lprule$ does not explicitly appear, which is needed to take gradients.
Towards this, we rewrite the first occurrence of $\lpdftargsg$ in Eq.~\eqref{eq:KL} as
\begin{align}
  \lpdftargsg(\ltarg) &= \int_{\lfeat\in\lfeatspace} \fpdf{\lrtarg|\lrsel=1,\lrfeat}(\ltarg,\lfeat) \fpdf{\lrfeat|\lrsel=1}(\lfeat) d\lfeat \\
  \label{eq:targsg-pdf}            &= \int_{\lfeat\in\lfeatspace}\fpdf{\lrtarg|\lrsel=1,\lrfeat}(\ltarg,\lfeat)   
  \frac{\fpdf{\lrsel=1|\lrfeat}(\lfeat)\fpdf\lrfeat(\lfeat)}{\fprb{\lrsel=1}}d \lfeat \,, \; \;
\end{align}
by using the rules of marginal probability resp.~Bayes' rule. We will first approximate Eq.~\eqref{eq:targsg-pdf}, and then subsequently show how we can estimate KL divergence of Eq.~\eqref{eq:KL} for its optimization.

To this end, we first note that the subgroup indicator $\lrsel$ takes two discrete values, indicating whether $\lfeat$ belongs to the subgroup.
The rule function $\learner(\lfeat)$ is deterministic in the limit of $\lpartemp\to 0$ as per Theorem \ref{eq:binning}.
We use this to partition the domain of integration $\lfeatspace$ into $\lfeatspacep \ldef \{\lfeat \in \lfeatspace| s_{t\rightarrow0}(\lfeat) = 1\}$ and $\lfeatspacem \ldef \{\lfeat \in \lfeatspace| s_{t\rightarrow0}(\lfeat) = 0\}$.
Under the following four assumptions we can bound the approximation of density $\lpdftargsg$ in Eq.~\eqref{eq:targsg-pdf}.
First, we assume that both $\fpdf\lrfeat$ and $\fpdf{\lrtarg|\lrsel=1,\lrfeat}$ are upper bounded by the finite constants $C_\lrfeat>0$ and $C_\lrtarg>0$, respectively. 
Secondly, we assume that in a subset $\lfeatspacems\subset\lfeatspacem$, where $s$ is neither zero or one, 
i.e. the yellow region in Figure \ref{fig:membership-box}, is negligable. Lastly, we assume that $\lfeatspacems$
covers almost all of the non-membership domain $\lfeatspacem$ and the probabilitiy mass is area is also negligable.
Formally,
\begin{align}
\fpdf\lrfeat(\lfeat)\leq C_\lrfeat,  \label{eq:ass-bnd-x} \\ 
\fpdf{\lrtarg|\lrsel=1,\lrfeat}\leq C_\lrtarg,\label{eq:ass-bnd-y}\\	
  \int_{\lfeat\in\lfeatspacems}\lpdfsel(\lfeat)d\lfeat\leq\epsilon_1\;,  \label{eq:ass-memb} \\
  \int_{\lfeat\in\lfeatspacem\setminus\lfeatspacems}\fpdf\lrfeat(\lfeat)d\lfeat<\epsilon_2\;. \label{eq:ass-align}
\end{align}

\begin{restatable}{theorem}{theoremapx}
  \label{thm:approx}
  When Eqs.~\eqref{eq:ass-bnd-x}, \eqref{eq:ass-bnd-y}, \eqref{eq:ass-memb}, and \eqref{eq:ass-align} hold, it is
  \begin{align}
    \lpdftargsg(\ltarg) - \int_{\lfeat\in\lfeatspacep}\fpdf{\lrtarg|\lrsel=1,\lrfeat}(\ltarg,\lfeat)\leq\frac{C_\lrtarg(\epsilon_2+C_\lrfeat\epsilon_1)}{\fprb{\lrsel=1}}d\lfeat\,.
  \end{align}
  Further, during learning, this bound becomes tighter until it asymptotically vanishes, assuming a decreasing annealing schedule for the temperature parameter.
\end{restatable}
We postpone the proof to Appendix~\ref{sec:approx}.

Using the same assumptions, we further approximate for all $\lfeat\in\lfeatspacep$ the target property conditional 
\begin{align}
    \fpdf{\lrtarg|\lrfeat}(\ltarg,\lfeat) &=  \fpdf{\lrtarg|\lrsel=1, \lrfeat}(\ltarg,\lfeat) \fpdf{\lrsel=1|\lrfeat}(\lfeat)  \\
                                          &+ \fpdf{\lrtarg|\lrsel=0, \lrfeat}(\ltarg,\lfeat) \fpdf{\lrsel=0|\lrfeat}(\lfeat) \\
                                          &\approx \fpdf{\lrtarg|\lrsel=1,\lrfeat}(\ltarg,\lfeat),
\end{align}
which allows us to approximate the subgroup-conditional target distribution from Eq.~\eqref{eq:targsg-pdf} as 
\begin{align}
  \lpdftargsg(\ltarg) &= \int_{\lfeat \in \lfeatspacep} \fpdf{\lrtarg|\lrsel=1,\lrfeat}(\ltarg,\lfeat)\frac{\fpdf{\lrsel=1|\lrfeat}(\lfeat)\fpdf{\lrfeat}(\lfeat)}{\fprb{\lrsel=1}} dx \\
  &\approx \int_{\lfeat \in \lfeatspacep} \fpdf{\lrtarg,\lrfeat}(\ltarg,\lfeat)\frac{\fpdf{\lrsel=1|\lrfeat}(\lfeat)}{\fprb{\lrsel=1}} dx\;.
  \label{eq:targ-approx}
\end{align}
Finally, we replace Eq.~\eqref{eq:targ-approx} into Eq.~\eqref{eq:KL} to obtain our final approximation
\begin{align}
    \label{eq:full-KL}
    &\xkldiv \lcdftargsg \lcdftargsg \approx \\ &\int_{\ltarg \in \ltargspace} \int_{\lfeat \in \lfeatspacep} \fpdf{\lrtarg,\lrfeat}(\ltarg,\lfeat)\frac{\fpdf{\lrsel=1|\lrfeat}(\lfeat)}{\fprb{\lrsel=1}} dx \log\left(\frac{\lpdftargsg(\ltarg)}{\lpdftarg(\ltarg)}\right) d\ltarg \;.
\end{align}
From this point onward we can use the standard Monte Carlo estimation of this integral, which gives
\begin{equation}
    \label{eq:score}
    \xkldiv\lcdftargsg\lcdftarg\approx \frac{1}{\mean} \sum_{k=1}^\lnsamp \lprule(\lfeat^{(k)})\log\left(\frac{\lpdftargsg(\ltarg^{(k)})}{\lpdftarg(\ltarg^{(k)})}\right) \;,
\end{equation}
where $\lpdftarg$ and $\lpdftargsg$ stand for the models trained from the normalising flows, $\lprule$ is our subgroup membership model (see Sec.~\ref{sec:rule-induction}) and $\mean$ is estimated as $\frac1\lnsamp\sum_{i=1}^\lnsamp \lprule(\lfeat^{(i)})$.
Our approximation is directly computable given the density estimates $\lpdftarg$ and $\lpdftargsg$ from the normalizing flows.
Crucially this allows us to update the subgroups rule $\lprule$ as to maximize its exceptionality/KL-Divergence.
Finally, we can now deal with some fine tuning of the objective to discover both representative and diverse subgroups.

\subsection{Rule Generality and Diversity}
The KL-divergence measures dissimilarity between two distributions, but naively maximising it has a drawback.
We could easily craft a small subgroup consisting of the most deviating sample on its own, defined by a rule with a very narrow scope and relatively low value.
Thus, we employ a common technique in subgroup discovery~\citep{boley_identifying_2017} in order to steer the results towards larger subgroups: we multiply the statistic of dissimilarity with the size of the subgroup $\mean^\lweisize$. The power $\lweisize$ tunes the trade-off in the importance of subgroup exceptionality and size. 

As we will show in our experiments, traditional subgroup discovery approaches often find a set of nearly identical subgroups.
In the typical top-$k$ scheme, the best scored subgroups are often slight variations of the same rule.
To encourage \ourmethod to learn subgroups with diverse distributions, we introduce a \fdef{regularizer},
where we add the KL-divergence of the current subgroup $S$ to the previously found subgroups $S_k$ to our objective.
Hence, summarising all the above,
we obtain as our final objective our variant of the \fdef{size-corrected KL}~\citep{leeuwen_non-redundant_2011} $\wkldiv\lcdftargsg\lcdftarg$:
\begin{align}
  &\mean^\lweisize \expandafter \hat\xkldiv\lcdftargsg\lcdftarg
  + \lambda \sum_{j=1}^j \expandafter \hat\xkldiv\lcdftargsg{\lcdf_{\lrtarg|S_j=1}}\;.
  \label{eq:weighted-KL}
\end{align}
The parameter $\lambda$ controls the strength of the regularizer. %

\subsection{Full Model}
\label{sec:architecture}
In the previous sections, we detailed our rule learning architecture with differentiable thresholding and aggregation (Sec.~\ref{sec:rule-induction}).
We described how to use Neural Spline Flows to obtain non-parametric density estimates (Sec.~\ref{sec:flows}), and finally derived Objective \eqref{eq:weighted-KL}, a 
size aware Kullback-Leibler Divergence that allows us to optimize our rule function $s(\lfeat)$ with gradient descent.
Together these make up the components of our \ourmethod architecture for subgroup discovery with normalising flows.
Given a dataset $\{(\lfeat^{(k)},y^{(k)})\}_{k=1}^N\sim\fprb{\lrfeat,\lrtarg}$,
\ourmethod undergoes the following three steps for each sample $(\lfeat^{(k)},y^{(k)})$:
\begin{tidyenum}
  \item \emph{Feature Thresholding}: All continuous features $\lfeat_i^{(k)}$ are thresholded with learned parameters $\llb_i$ and $\lub_i$ 
  using the soft-binning from Eq.~\eqref{eq:soft-predicate}.
  Thereby we obtain a predicate vector $\pred(\lfeat^{(k)};\llb,\lub,\lpartemp) \in [0,1]^p$.
\item \emph{Subgroup Rule}: We employ weights $a_i$ to combine the individual predicates $\pred_i(\lfeat^{(k)}_i)$ into a conjunction $s(\lfeat;\llb,\lub,a,\lpartemp)$.
This rule represents the probability of subgroup membership $\lprb(\lrsel=1|X=\lfeat^{(k)})$.
\item \emph{Distribution Exceptionality}: We estimate the likelihood of $\fpdf{\lrtarg}(y^{(k)})$ and $\fpdf{\lrtarg|S=1}(y^{(k)})$ with two 
separately fitted normalising flow models. Then, according to Objective \eqref{eq:weighted-KL}, we can estimate the KL-Divergence between the current subgroup
and the general distribution.
\end{tidyenum}

By repeating the aforementioned steps over all samples $(\lfeat^{(k)},y^{(k)})$ and summing up the results, Objective \eqref{eq:weighted-KL} gives us a differentiable
estimate of the KL-Divergence in regards to the subgroup rule $s(\lfeat)$.
We optimize $s(\lfeat)$ using standard gradient descent techniques with the Adam optimizer \citep{kingma:15:adam}.
After the subgroup rule has been updated, we again update the normalising flow of the subgroups density as described in Sec.~\ref{sec:flows}, and 
repeat this process for a user-specified amount of epochs.
During the training, we gradually decrease the temperature $t$ by a pre-determined schedule to obtain increasingly crisp subgroup assignments.
Finally, at the last epoch, the discovered subgroup is then the output of the subgroup rule $s(\lfeat)$.
We provide a diagram overviewing and the pseudo-code for \ourmethod in the Appendix \ref{ap:algo}.


%
\section{Related Work}
\label{sec:related}

\paragraph{Subgroup Discovery.}
Traditional approaches for subgroup discovery~\citep{klosgen:96:explora} can be split based on type of search and on exceptionality measures. 
Subgroup discovery is NP-hard~\citep{boley_non-redundant_2009} and hence most proposals resort to greedy heuristics~\citep{duivesteijn_exceptional_2016,atzmueller_subgroup_2015} without guarantees. Branch-and-bound based algorithms~\citep{boley_identifying_2017,kalofolias_discovering_2019} permit results with guarantees for some exceptionality measures, but generally do not scale beyond tens of features~\citep{atzmueller_sd-map_2006}. 
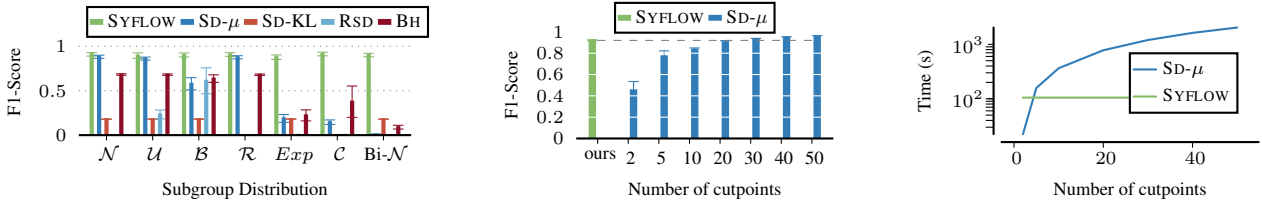
\begin{figure*}[!ht]
    \begin{subfigure}[t]{0.35\linewidth}
        \centering
        \begin{tikzpicture}
            \begin{axis}[
                pretty ybar,
                pretty enlargexlimits,
                cycle list name = prcl-ybar,
                symbolic x coords = {$\mathcal{N}$,$\mathcal{U}$,$\mathcal{B}$,$\mathcal{R}$,$Exp$,$\mathcal{C}$,Bi-$\mathcal{N}$},
                bar width = 0.08em,
                height = 3cm,
                ymajorgrids, 
		        major grid style   = {gray,dotted},
                width=6cm,
                ylabel = {F1-Score},
                xlabel = {Subgroup Distribution},
                pretty labelshift,
                legend columns = 7,
                legend entries = {\ourmethod, \sd,\sdkl,\rsd,\bh},
                legend style = {xshift=-1.6cm,yshift=0.3cm, font=\scriptsize},
                ymax=1.2,
                ymin=0.0
                ]
                \addplot+[error bars/.cd, 
                y fixed,
                y dir=both, 
                y explicit] table[x={Distribution},y={F1},col sep=comma, y error plus expr=\thisrow{F1-STD}, y error minus expr=\thisrow{F1-STD}] {expres/target_distribution/syflow.csv};

                \addplot+[error bars/.cd, 
                y fixed,
                y dir=both, 
                y explicit] table[x={Distribution},y={F1},col sep=comma,  y error plus expr=\thisrow{F1-STD}, y error minus expr=\thisrow{F1-STD}] {expres/target_distribution/sd-mean.csv}; 

                 \addplot+[error bars/.cd, 
                 y fixed,
                 y dir=both, 
                 y explicit] table[x={Distribution},y={F1},col sep=comma,  y error plus expr=\thisrow{F1-STD}, y error minus expr=\thisrow{F1-STD}] {expres/target_distribution/sd-kl.csv}; 
                \addplot+[error bars/.cd, 
                y fixed,
                y dir=both, 
                y explicit] table[x={Distribution},y={F1},col sep=comma,  y error plus expr=\thisrow{F1-STD}, y error minus expr=\thisrow{F1-STD}] {expres/target_distribution/rsd.csv}; 
                \addplot+[error bars/.cd, 
                y fixed,
                y dir=both, 
                y explicit] table[x={Distribution},y={F1},col sep=comma,  y error plus expr=\thisrow{F1-STD}, y error minus expr=\thisrow{F1-STD}] {expres/target_distribution/bh.csv}; 
            \end{axis}
    \end{tikzpicture}
    \subcaption{F1-scores of recovering a planted subgroup for different target distributions, higher is better.}
    \label{fig:noise}
    \end{subfigure}%
    \hfill
    \begin{subfigure}[t]{0.3\linewidth}
        \centering
        \begin{tikzpicture}
        \begin{axis}[
            pretty ybar,
            pretty enlargexlimits, %
            cycle list name = prcl-ybar,
            symbolic x coords = {1,2,5,10,20,30,40,50},
            ymax = 1,
            ymin=0,
            pretty labelshift,
            xtick={1,2,5,10,20,30,40,50},
            xticklabels = {ours,2,5,10,20,30,40,50},
            bar width = 0.2em,
            width = \linewidth,
            height = 3cm,
            width = 5cm,
            ylabel = {F1-Score},
            xlabel = {Number of cutpoints},
            legend entries = {\ourmethod,\sd},
            legend style={xshift=-1.5cm,yshift=0.4cm, font=\scriptsize},
            legend columns = 2
            ]
            \addplot+[dollarbill,error bars/.cd, 
            y fixed,
            y dir=both, 
            y explicit
            ]
            coordinates {
            (1,0.92) +=(0,0.0055) -= (0,0.0055)
            };
            \addplot+[error bars/.cd, 
            y fixed,
            y dir=both, 
            y explicit] table[x=N-bins,y=F1,col sep=comma,y error plus expr=\thisrow{F1-STD}]{expres/cutpoints/sd-mean.csv};
            \draw [example thresh] (1,0.92) -- (50,0.92);
        \end{axis}
        \label{fig:cp}
    \end{tikzpicture}
    \caption{F1-scores for \ourmethod and \sd, higher is better.}
    \label{fig:cp}
    \end{subfigure}%
    \hfill
    \begin{subfigure}[t]{0.3\linewidth}
        \centering
        \begin{tikzpicture}
            \usetikzlibrary{calc}
            \begin{semilogyaxis}[ 
                pretty line,
                cycle list name = prcl-line,
                legend entries = {\sd,\ourmethod},
                legend style={xshift=0cm,yshift=-0.8cm, font=\scriptsize},
                pretty enlargexlimits,
                pretty labelshift,
                height=3cm,
                width=5cm,
                ylabel={Time (s)},
                xlabel={Number of cutpoints},
                ymin=0,
                legend columns = 1,
                xmin=2,
                xmax=50,
                legend style={xshift=-0.2cm,yshift=0.5cm},
                tick label style={/pgf/number format/fixed}
            ] %
            
            \addplot+[b2] table[x={N-bins},y={Runtime},col sep=comma]{expres/cutpoints/sd-mean.csv};
            \draw [dollarbill] (2,105) -- (50,105);
            \addplot[dollarbill] coordinates{ (2,105)    };
            \end{semilogyaxis}
          \end{tikzpicture}
          \caption{Runtime of \ourmethod and \sd, lower is better.}
          \label{fig:cpruntime}
    \end{subfigure}
    \caption{\textit{Subgroup Predictive Accuracy.} \textbf{(a)} Method comparison in terms of F1-score recovering subgroups in synthetic data. 
    \textbf{(b)} Across different distributions: \ourmethod outperforms the competition on distributions with higher order moments. 
	\textbf{(c)} With increasing number of cutpoints, \sd matches \ourmethod accuracy around 40 bins, but needs 10$\times$ more time.}
    \label{fig:exp1}
\end{figure*}


Most proposals for subgroup discovery assess exceptionality by comparing the conditional and marginal distributions of a single univariate target \citep{song_subgroup_2016,helal2016subgroup}. 
Basic measures of exceptionality include mean-shift~\citep{grosskreutz_subgroup_2009} for continuous, and weighted relative accuracy~\citep{song_subgroup_2016} for discrete targets, but there exists a wide range of proposals for many data types~\citep{kalofolias_discovering_2019}. Most, however, make strong assumptions about the distribution of the target such as normal~\citep{friedman:1999:bump,lavrac_subgroup_2004}, binomial or $\chi^2$~\citep{grosskreutz_subgroup_2009}. 

\citet{duivesteijn_exceptional_2016} generalize subgroup discovery to multivariate targets by measuring exceptionality based on the difference between models just trained on the subgroup versus on all of the data. Due to computational costs, only simple models can be used, leading to a compromise in performance.
Similarly, \citet{izzo2023data} search for subgroups that can be described by a linear model.
\citet{proencca:2022:robust} instead measure exceptionality using a proxy of KL-divergence based on the Minimum Description Length (MDL) principle. In contrast, \ourmethod  optimizes KL-divergence directly, can employ any type of normalizing flow, and is equally suited for uni-/multi-variate targets.
\paragraph{Differentiable Rule Induction}
Classical rule induction methods aim to find rules of the form ``$\textbf{ If } X_1 = 1 \wedge X_5=1 \textbf{ Then } Y=0$'' through expensive combinatorial optimization. 
Recently, highly scalable differentiable rule induction methods were proposed based on differentiable analogues of logical connectives, e.g. conjunctions and disjunctions~\cite{fischer2021differentiable, wang_transparent_2020} that permit extracting crisp logical rules after training.
Most work in this direction focuses on learning a global classifier \citep{yang:18:deeptree,qiao2021learning,wang2021scalable, dierckx2023rl} as opposed to our goal of learning concise rule that identify exceptional subgroups. In this sense most related is \citet{walter2024finding}, who propose a neural architecture to find conjunctions of binary features that are over resp. under-expressed for a particular label. In contrast, \ourmethod considers continuous features, and is not constrained to a type or number of target variables.


\section{Experiments}
\label{sec:experiments}

We evaluate \ourmethod against four state-of-the-art methods on synthetic and real-world data.
We compare against Bump Hunting~\citep[\bh,][]{friedman:1999:bump}, subgroup discovery using mean-shift~\citep[\sd,][]{lemmerich:2018:pysubgroup}, subgroup
discovery using KL-divergence (\sdkl), and Robust Subgroup Discovery \citep[\rsd,][]{proencca:2022:robust}.
We give the hyperparameters in Appendix \ref{ap:hyperparameters} and provide the code in the Supplementary Material.

\subsection{Synthetic Data}\label{sec:synthetic}
To evaluate on datasets with known ground truth we consider synthetic data.
We begin by generating $\lnfeat$ feature variables $\lrfeat_i$ and the target variable $\lrtarg$ from a uniform distribution $\mathcal{U}(0,1)$
to create a dataset of $n=20\,000$ samples.
Within this dataset, we plant a rule $\sigma(x)=\land_i^c \xpred {\lfeatel_i} {\flb i} {\fub i}$ of $c$ conditions.
The hypercube described by the rule is set to have a volume of 0.1 (10\% of population).
For the samples within the planted subgroup, we re-sample the target variable $Y$ using a separate distribution $\lprb(\lrtarg|\lrsel=1)$.
We run each experiment five times and report the average.

    \begin{figure}[!b]
        \begin{subfigure}[t]{0.49\linewidth}
            \centering
            \begin{tikzpicture}
                \usetikzlibrary{calc}
                \begin{axis}[ 
                    pretty line,
                    cycle list name = prcl-line,
                    legend style={yshift=0.7cm,font=\scriptsize,xshift=3.4cm},
                    width=\linewidth,
                    height=3cm,
                    ylabel={F1-score},
                    xlabel={Number of features $\lnfeat$},
                    pretty labelshift,
                    ymax=1.0,
                    xmode=log,
                    legend columns = 5,
                    legend entries = {\ourmethod, \sd,\sdkl,\rsd,\bh},
                    tick label style={/pgf/number format/fixed}
                ]
                \foreach \x in {syflow, sd-mean, sd-kl,rsd, bh}{
                    \addplot+[error bars/.cd, 
                    y fixed,
                    y dir=both, 
                    y explicit] table[x={Dimension},y={F1},col sep=comma]{expres/scaling/\x.csv};
                }      
                \end{axis}
              \end{tikzpicture}
              \subcaption{F1 under increasingly more features, higher is better.}
              \label{fig:scalability-f1}
        \end{subfigure}%
        \hfill
        \begin{subfigure}[t]{0.49\linewidth}
            \centering
            \begin{tikzpicture}
                \usetikzlibrary{calc}
                \begin{axis}[ 
                    pretty line,
                    cycle list name = prcl-line,
                    pretty labelshift,
                    legend style={xshift=1cm,yshift=0.3cm},
                    width=\linewidth,
                    height=3cm,
                    ylabel={Runtime (s)},
                    xlabel={Number of features $\lnfeat$},
                    xmode=log,
                    tick label style={/pgf/number format/fixed},
                    yticklabels =  {0,0,2k,4k,6k,8k}
                ] 
                \foreach \x in {syflow, sd-mean, sd-kl, rsd, bh}{
                    \addplot+[error bars/.cd, 
                    y fixed,
                    y dir=both, 
                    y explicit] table[x={Dimension},y={Runtime},col sep=comma]{expres/scaling/\x.csv};
                }      
                \end{axis}
              \end{tikzpicture}
              \subcaption{Runtime under increasingly more features $\lnfeat$, lower is better. }
              \label{fig:scalability-runtime}
        \end{subfigure}%
        \caption{Scalability of \ourmethod and baselines.}
        \label{fig:scalability}
\end{figure}
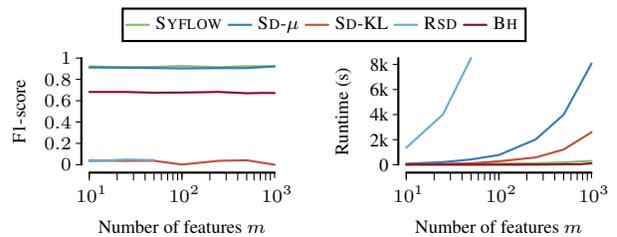

\paragraph{Target Distribution}

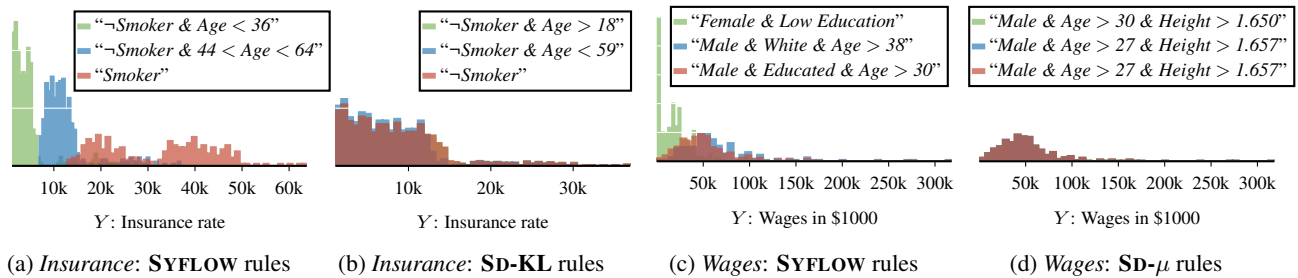
\begin{figure*}[!t]
    \begin{subfigure}{0.25\textwidth}
    \centering
    \begin{tikzpicture}
            \begin{axis}[
                pretty ybar,
                cycle list name = prcl-ybar,
                height = 3.5cm,
                width=5.5cm,
                xlabel = {$\lrtarg$: Insurance rate},
                ytick={},
                yticklabels={},
                ymax=0.00025,
                ymin=0,
                scaled x ticks=false,
                scaled y ticks=false,
                xtick={10000,20000,...,60000},
                xticklabels={10k,20k,30k,40k,50k,60k},
                legend columns = 1,
                legend entries = {
                \fdesc{¬Smoker \& Age $<$ 36}, 
                \fdesc{¬Smoker \& 44 $<$ Age $<$ 64}, 
                \fdesc{Smoker},
                },
                legend style = {xshift=-1cm,yshift=0.15cm, font=\scriptsize}
            ]
            \foreach \x in {2,1,3}{
        \addplot +[
            hist={
                bins=50,
                density
            },
            opacity=0.7
        ] table [y index=0] {expres/insurance/syflow-subgroup\x.txt};
        }
        \end{axis}
    \end{tikzpicture}
    \subcaption{\emph{Insurance}: \textbf{\ourmethod} rules}
    \label{fig:insurance-syflow}
\end{subfigure}%
\hfill
\begin{subfigure}{0.25\textwidth} 
    \centering
    \begin{tikzpicture}
        \begin{axis}[
            pretty ybar,
            cycle list name = prcl-ybar,
            height = 3.5cm,
            width=5.5cm,
            xlabel = {$\lrtarg$: Insurance rate},
            ytick={},
            yticklabels={},
            ymax=0.00025,
            ymin=0,
            scaled x ticks=false,
            scaled y ticks=false,
            xtick={10000,20000,...,60000},
            xticklabels={10k,20k,30k,40k,50k,60k},
            legend columns = 1,
            legend entries = {
            \fdesc{¬Smoker \& Age $>$ 18},
            \fdesc{¬Smoker \& Age $<$ 59}, 
            \fdesc{¬Smoker},
            },
            legend style = {xshift=-1cm,yshift=0.15cm, font=\scriptsize}
        ]
        \foreach \x in {2,3,1}{
    \addplot +[
        hist={
            bins=50,
            density
        },
        opacity=0.7
    ] table [y index=0] {expres/insurance/sd-kl-subgroup\x.txt};
    }
    \end{axis}
\end{tikzpicture}
    \subcaption{\emph{Insurance}: \textbf{\sdkl} rules}
    \label{fig:insurance-sd}
\end{subfigure}%
\hfill
\begin{subfigure}{0.25\textwidth}
    \begin{tikzpicture}
        \begin{axis}[
            pretty ybar,
            cycle list name = prcl-ybar,
            height = 3.5cm,
            width=5.5cm,
            xlabel = {$\lrtarg$: Wages in \$1000},
            ytick={},
            yticklabels={},
            ymax=0.0001,
            ymin=0,
            scaled x ticks=false,
            scaled y ticks=false,
            xtick={50000,100000,...,300000},
            xticklabels={50k,100k,150k,200k,250k,300k},
            legend columns = 1,
            legend style = {xshift=-1.5cm,yshift=0.15cm, font=\scriptsize,cells={align=center}},
            legend entries = {\fdesc{Female \& Low Education},
                \fdesc{Male \& White \& Age $>$ 38}, 
                \fdesc{Male \& Educated \& Age $>$ 30}
            },
        ]
        \foreach \x in {4,3,1}{
    \addplot +[
        hist={
            bins=50,
            density
        },
        opacity=0.7
    ] table [y index=0] {expres/wages/syflow-subgroup\x.txt};
    }
    \end{axis}
\end{tikzpicture}
\subcaption{\emph{Wages}: \textbf{\ourmethod} rules}
\label{fig:wages-syflow}
\end{subfigure}%
\hfill
\begin{subfigure}{0.25\textwidth} 
\begin{tikzpicture}
    \begin{axis}[
        pretty ybar,
        cycle list name = prcl-ybar,
        height = 3.5cm,
        width=5.5cm,
        xlabel = {$\lrtarg$: Wages in \$1000},
        ytick={},
        yticklabels={},
        ymin=0,
        ymax=0.0001,
        scaled x ticks=false,
        scaled y ticks=false,
        xtick={50000,100000,...,300000},
        xticklabels={50k,100k,150k,200k,250k,300k},
        legend columns = 1,
        legend entries = {
        \fdesc{Male \& Age $>$ 30 \& Height $>$ 1.650},
        \fdesc{Male \& Age $>$ 27 \& Height $>$ 1.657}, 
        \fdesc{Male \& Age $>$ 27 \& Height $>$ 1.657}
        },
        legend style = {xshift=-1.5cm,yshift=0.15cm, font=\scriptsize,cells={align=left}}
    ]
    \foreach \x in {1,2,3}{
\addplot +[
    hist={
        bins=50,
        density
    },
    opacity=0.7
] table [y index=0] {expres/wages/sd-mean-subgroup\x.txt};
}
\end{axis}
\end{tikzpicture}
\subcaption{\emph{Wages}: \textbf{\sd} rules}
\label{fig:wages-sd}
\end{subfigure}%
    \caption{Subgroups learned on the \textit{Insurance} and \emph{Wages} datasets.
    Only \ourmethod learns {diverse} and {exceptional} subgroups.}
    \label{fig:realworld}
\end{figure*}

First, we assess for all methods their accuracy in recovering the planted subgroup for different distributions of the target property $\lrtarg$.
To this end, we vary $\lprb(\lrtarg|\lrsel=1)$ to be respectively a normal distribution $\mathcal{N}(1.5,0.5)$, uniform distribution $\mathcal{U}(0.5,1.5)$, exponential distribution $Exp(0.5)$, rayleigh distribution $\mathcal{R}(2)$, cauchy distribution $\mathcal{C}(0,1)$, beta distribution $\mathcal{B}(0.2,0.2)$,  and a balanced mixture distribution of two gaussians (Bi-$\mathcal{N}$) $\mathcal{N}(-1.5,0.5)$ and $\mathcal{N}(1.5,0.5)$. The distribution are shown in Figure \ref{appfig:target} in Appendix \ref{app:target}.
The number of features $\lnfeat=10$ and complexity $c=4$ remains fixed.

For each method, we compute the F1-score between the ground truth and discovered subgroup labels (i.e. $S=1$).
We present our results in Figure \ref{fig:noise}; We see that for distributions that are well characterized by their first moment, i.e.~the uniform and normal distribution, \ourmethod, \sd and \bh are all able to recover the planted subgroup.
On the exponential, Rayleigh and bi-modal distributions, only \ourmethod finds the planted subgroup.
In general, \ourmethod reliably recovers subgroups independent of the underlying target distribution.

\paragraph{Thresholding}
Next, we study the efficacy of the differentiable feature thresholding. We generate data as before, considering only Normal distributions for the target variable $\lrtarg$, and setting $\lnfeat=50$. We compare the accuracy of \ourmethod against \sd, whilst gradually increasing the amount of bins per feature in the pre-processing for \sd.

As we can see in Figure~\ref{fig:cp}, as the number of cutpoints increases, the F1-score of \sd improves. 
Under 20 cutpoints \sd performs much worse than \ourmethod, while for more cutpoints it slightly outperforms it in terms of accuracy. At the same time, as the number of cutpoints increases, \sd runtime increases rapidly (Figure~\ref{fig:cpruntime}), requiring an order of magnitude more runtime to perform on par with \ourmethod.  

\paragraph{Scalability}
In the final experiment on synthetic data we assess performance and runtime when varying the number of features.
We allow each method up to 24 hours. We present our results in Figure~\ref{fig:scalability}.

We observe that all methods perform remarkably stable in terms of accuracy, as well as that \ourmethod is significantly faster than all of its competitors. As a continuous optimization based method, \ourmethod avoids the typical combinatorial explosion in runtime,
and additionally takes advantage of GPU acceleration. In comparison, \sd takes 50 times longer for $1\,000$ features, whereas \rsd does not finish within 24 hours for more than $100$ features.

In Appendix \ref{ap:rulecomplexity}, we additionally show that \ourmethod also scales well with increasing rule complexity, i.e.~the number of predicates, whereas its competitors with fixed discretization struggle.

\begin{table*}[ht]
    \centering
    \fontsize{8}{10}\selectfont
    \caption{Quantitative results of exceptionality of the subgroups discovered by resp. \ourmethod (ours), \sdkl, \sd, \rsd, and \bh,
     as measured KL-Divergence (\DKL, higher is better), Bhattacharyya coefficient (\BC, lower is better), and absolute mean distance (\AMD, higher is better). 
    }
    \begin{tabular}{l rrrrr rrrrr rrrrr}
        \toprule
        & \multicolumn{5}{c}{\DKL} & \multicolumn{5}{c}{\BC} & \multicolumn{5}{c}{\AMD} \\
         \cmidrule(r){2-6} \cmidrule(lr){7-11} \cmidrule(lr){12-16}
         & \emph{ours} & \sdkl & \sd & \rsd & \bh & \emph{ours} & \sdkl & \sd & \rsd & \bh & \emph{ours} & \sdkl & \sd & \rsd & \bh \\
         \midrule
         Abalone & \textbf{0.14} & 0.02 & 0.12 & 0 & 0.05 & \textbf{0.66} & 0.99 & 0.93 & 1 & 0.87 & 0.73 & 0.25 & \textbf{0.84} & 0 & 0.16 \\
        Airquality & 0.22 & 0.22 & \textbf{0.24} & 0 & 0.0 & \textbf{0.62} & 0.86 & 0.79 & 1 & 1.0 & 0.37 & \textbf{0.53} & 0.49 & 0 & 0.0 \\
        Automobile & 0.22 & 0.24 & 0.23 & \textbf{0.26} & 0.21 & 0.64 & 0.85 & 0.79 & 0.64 & \textbf{0.6} & 1838 & \textbf{2807} & 2683 & 2218 & 2475 \\
        Bike & \textbf{0.17} & 0.1 & 0.15 & 0.17 & 0.13 & \textbf{0.64} & 0.95 & 0.9 & 0.67 & 0.73 & 584 & 570 & \textbf{630} & 431 & 622 \\
        California & \textbf{0.13} & 0.06 & 0.11 & 0 & 0.0 & \textbf{0.72} & 0.97 & 0.93 & 1 & 1.0 & 0.25 & 0.3 & \textbf{0.32} & 0 & 0.0 \\
        Insurance & \textbf{0.27} & 0.13 & 0.26 & 0 & 0.19 & 0.55 & 0.93 & \textbf{0.52} & 1 & 0.84 & 3845 & \textbf{3973} & 3845 & 0 & 1518 \\
        Mpg & \textbf{0.27} & 0.26 & 0.24 & 0.21 & 0.24 & 0.57 & 0.76 & 0.8 & \textbf{0.47} & 0.61 & \textbf{2.99} & 2.85 & 2.96 & 1.66 & 2.79 \\
        Student & 0.08 & 0.03 & 0.08 & \textbf{0.09} & 0.04 & 0.86 & 0.99 & 0.94 & \textbf{0.71} & 0.97 & 0.46 & 0.52 & \textbf{0.69} & 0.47 & 0.45 \\
        Wages & \textbf{0.1} & 0.02 & 0.1 & 0 & 0.03 & \textbf{0.81} & 0.99 & 0.9 & 1 & 0.99 & \textbf{6043} & 2994 & 5916 & 0 & 5149 \\
        Wine & \textbf{0.08} & 0.0 & 0.06 & 0 & 0.01 & \textbf{0.89} & 1.0 & 0.97 & 1 & 0.97 & 0.17 & 0.04 & \textbf{0.19} & 0 & 0.04 \\
        \midrule
        Avg.~rank & \textbf{1.5} & 3.5 & 2.1 & 3.5 & 3.6 & \textbf{1.4} & 4.0 & 2.8 & 3.3 & 2.9 & 2.6 & 2.4 & \textbf{1.5} & 4.5 & 3.6 \\
        \bottomrule
    \end{tabular}
    \label{tab:real-world}
\end{table*}

\subsection{Real World Data}
\label{sec:realworld}
We now turn to real-world data, where we evaluate on regression datasets from the UCI-Machine Learning Repository.\!\footnote{\url{https://archive.ics.uci.edu}.}
They provide a variety of target distributions and feature spaces to search, and allow us qualitatively inspect the learned rules.
On each dataset, we let each method return $5$ subgroups. As the ground truth is unknown,
we asses the subgroup quality in terms of KL-divergence (\DKL), the distribution overlap or Bhattacharyya Coefficient (\BC), 
and by absolute difference in mean (\AMD). As \DKL and \AMD are strongly influenced by the size \mean of the subgroup, we correct for this. 
For a formal definition we refer to Appendix \ref{app-sec:metrics}.
In Table \ref{tab:real-world} we report, per metric, the scores of the best subgroup each method found.

Across the board, \ourmethod reliably finds exceptional subgroups, and is either the best or close to the best method when it comes to the distribution based exceptionality measures (\DKL and \BC). \sd optimizes for mean difference, and hence it is not surprising that it outperforms \ourmethod on this metric, but that \ourmethod nevertheless comes so close in terms of scores is a very positive result indeed.

For the \emph{Insurance} dataset it is inconclusive which method learns the most exceptional subgroups; \ourmethod returns the best subgroup in terms of \DKL, \sdkl finds the subgroup with the largest mean difference.  
For further analysis, we plot the best three subgroups found by \ourmethod and \sdkl in Figure \ref{fig:insurance-syflow} and \ref{fig:insurance-sd}.
The specific rules that \ourmethod finds are succinct and informative (\fdesc{¬Smoker \& Age $<$ 36}), and represent a diverse 
set of subgroups (low, medium and high premiums).
In contrast, \sdkl finds highly redundant rules, e.g.~\fdesc{¬Smoker} and \fdesc{¬Smoker \& Age $>$ 18}, that all describe 
the same target distribution.

Generally, on all datasets where the target variable is exponentially distributed (\emph{Wages}, \emph{Insurance}, \emph{California}), 
\ourmethod performs well in any measured metric, which matches the results seen on the exponential synthetic data (Sec.~\ref{sec:synthetic}).
Finally, in Fig.~\ref{fig:wages-syflow},\ref{fig:wages-sd}, we plot the learned subgroups on the \emph{Wages} dataset mentioned in the introduction.
Again, \ourmethod finds diverse subgroups of disadvantaged demographics (\fdesc{Female \& Low Education}) and advantaged groups (\fdesc{Male \& White \& Age $>$ 38}),
whilst its competitor find either much less exceptional subgroups (\rsd,\bh) or variants of the same subgroup (\sd,\sdkl).

Overall, \ourmethod shows versatility in finding exceptional subgroups across a variety of datasets and metrics.
The discovered rules are succinct and diverse, can scale to large datasets, and are robust to the underlying target distribution.
In the following, we will use \ourmethod on a specific application in materials science.

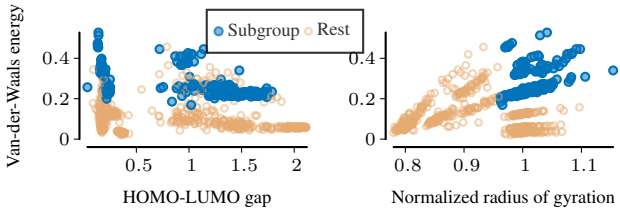
\begin{figure}[!b]
    \begin{subfigure}{0.45\linewidth}
        \centering
    \begin{tikzpicture}
        \begin{axis}[
            pretty scatter,
            example marks,
            pretty labelshift,
            legend entries = {Subgroup, Rest},
            legend style = {anchor = south,at={(0.9,0.8)}, opacity = 0.8},
            legend columns = 2,
            xlabel=HOMO-LUMO gap,
            ylabel=Van-der-Waals energy,  
            height=3cm,
            width=4.5cm]
            \addplot+[
            scatter src = explicit symbolic,              
            scatter/classes = {
                b={mark=*, frenchblue},
                a={mark=o, draw=fawn, mark size=1.25,opacity=0.5}
            }, only marks]
            table[x = Homolumo,y = vdw,meta = subgroup4, col sep=comma] {expres/gold/joint-subgroup.csv};
        \end{axis}
    \end{tikzpicture}
    \label{fig:joint-vdw}
    \end{subfigure}
    \hfill
    \begin{subfigure}{0.45\linewidth}
    \centering
    \begin{tikzpicture}
        \begin{axis}[
            pretty scatter,
            example marks,
            pretty labelshift,
            legend columns = 2,
            xlabel=Normalized radius of gyration,
            ylabel=,   
            legend entries = {
            },
            height=3cm,
            width=4.5cm]
            \addplot+[
            scatter src = explicit symbolic,              
            scatter/classes = {
                b={mark=*, frenchblue},
                a={mark=o, draw=fawn, mark size=1.25,opacity=0.5}
            }, only marks]
            table[x = gyration,y = vdw,meta = subgroup4, col sep=comma] {expres/gold/joint-subgroup.csv};
        \end{axis}
    \end{tikzpicture}
    \label{fig:gyration}
\end{subfigure}
\caption{\emph{Gold Nano-Clusters.} \ourmethod discovers a subgroup of molecules that have an 
outstanding \emph{joint} distribution of Van-der-Waals energy \emph{and} HOMO-LUMO gap. }
    \label{fig:case-study}
\end{figure}

\subsection{Case Study: Materials Science}

Next, we consider a case study on materials science data~\citep{goldsmith2017uncovering}, an application where learning diverse exceptional subgroups has a clear scientific benefit. In particular, we consider a dataset of properties of gold-nano-clusters. These are key components in photo-voltaics, as well as in medical applications \cite{giljohann2020gold}. The key goal is to better understand which molecular configurations lead to materials with better properties in terms of absorbing photons or with
which cells they interact. 

We first focus on characterizing the HOMO-LUMO gap, the difference between the highest occupied and lowest unoccupied molecular orbit, which corresponds to the efficiency in energy absorption of the material for photons of specific frequencies. Gold nano-clusters are also increasingly used in medical applications, in which the HOMO-LUMO gap is crucial to determine their reactivity with other molecules.

We showed the subgroups that \ourmethod learns for the HOMO-LUMO gap in Figure \ref{fig:homo-lumo} in the Introduction.
The subgroups it finds are based on known ground-truth factors~\cite{goldsmith2017uncovering}, i.e.~odd number of atoms leads to low gaps, but also ones with more complex descriptors such as clusters with an even amount of atoms that are primarily made up of double bond connections.
\ourmethod provides \emph{out of the box} interpretable, exceptional and diverse subgroups on this non-trivial target quantity.

Next, we additionally consider the relative intramolecular Van der Waals energies as the target quantity of interest. These quantities are important to determine the strength of molecular binding, e.g. allow a compound to selectively interact with the intended target. 
We show the results in Figure \ref{fig:case-study}. It is easy to see that samples fitting subgroup rule $\sigma$ that \ourmethod learns is indeed exceptional with regard to the overall distribution. Moreover, when we analyze the rule in detail, it makes physical sense. Molecules of between 8 to 14 atoms with a low (close to 1) gyration tend to have a significantly higher Van der Waals energy than those that are more strongly gyrated (i.e. are less `flat')~\cite{goldsmith2017uncovering}.

\section{Conclusion}
\label{sec:conclusion}
We explored the problem of discovering a  diverse set of exceptional subgroups, where the distribution 
of the target within each subgroup differs significantly from the overall target distributions. Existing works
suffer from combinatorial explosion, can not deal with intricate real world distributions and heavily
depend on the pre-quantisation of the features.
To overcome these limitations, we propose  \ourmethod, where we formulate subgroup discovery as 
continuous optimization problem. We use normalizing flows to accurately learn the distribution of the target property,
enabling us to deal with arbitrary complicated distributions, and propose a differentiable rule learner,
which simultaneously learns the subgroup description and the corresponding discretization. 
We show on synthetic and real-world data, including a case study on gold nano-clusters,
that \ourmethod reliably discovers diverting subgroups, especially when the target distribution
is non-standard. On gold nano-clusters we find physically plausible subgroups.

A current limitation of all subgroup discovery methods, including \ourmethod, is that the description language of conjunctions of Boolean predicates may be to simple to describe complex subgroups for physical data. We are specifically interested in exploring how we can extend \ourmethod to perform symbolic regression \cite{ouyang2018sisso}.

\newpage
\section*{Broader Impact}
The main motivation of the work presented in this paper, is to develop a method,
which can assist practitioners to make new scientific discovery. For example,
new insights on gold-cluster can have potential impact on biomedical applications
and more efficient photovoltaic system.
However, when applied to sensitive census data, it is important to stress that \ourmethod
is based on correlations and is thus not capable to make a definite causal statement.

\bibliographystyle{icml2024}
\bibliography{bib/abbreviations,bib/bib-jilles,bib/bib-paper,bib/bib-goflow-janis}

\providecommand{\noopsort}[1]{}
\begin{thebibliography}{37}
\providecommand{\natexlab}[1]{#1}
\providecommand{\url}[1]{\texttt{#1}}
\expandafter\ifx\csname urlstyle\endcsname\relax
  \providecommand{\doi}[1]{doi: #1}\else
  \providecommand{\doi}{doi: \begingroup \urlstyle{rm}\Url}\fi

\bibitem[Atzmueller(2015{\natexlab{a}})]{atzmueller:15:subgroup}
Atzmueller, M.
\newblock Subgroup discovery.
\newblock \emph{Wiley Interdisciplinary Reviews: Data Mining and Knowledge
  Discovery}, 5\penalty0 (1):\penalty0 35--49, 2015{\natexlab{a}}.

\bibitem[Atzmueller(2015{\natexlab{b}})]{atzmueller_subgroup_2015}
Atzmueller, M.
\newblock Subgroup discovery.
\newblock \emph{Wiley Interdisciplinary Reviews: Data Mining and Knowledge
  Discovery}, pp.\  35--49, January 2015{\natexlab{b}}.

\bibitem[Atzmueller \& Puppe(2006)Atzmueller and Puppe]{atzmueller_sd-map_2006}
Atzmueller, M. and Puppe, F.
\newblock {{SD-Map}} \textendash{} {{A Fast Algorithm}} for {{Exhaustive
  Subgroup Discovery}}.
\newblock In \emph{Knowledge {{Discovery}} in {{Databases}}: {{PKDD}} 2006},
  pp.\  6--17. {Springer Berlin Heidelberg}, 2006.
\newblock ISBN 978-3-540-46048-0.

\bibitem[Boley \& Grosskreutz(2009)Boley and
  Grosskreutz]{boley_non-redundant_2009}
Boley, M. and Grosskreutz, H.
\newblock Non-redundant subgroup discovery using a closure system.
\newblock In \emph{Joint {{European Conference}} on {{Machine Learning}} and
  {{Knowledge Discovery}} in {{Databases}}}, pp.\  179--194. {Springer},
  {Springer}, 2009.

\bibitem[Boley et~al.(2017)Boley, Goldsmith, Ghiringhelli, and
  Vreeken]{boley_identifying_2017}
Boley, M., Goldsmith, B.~R., Ghiringhelli, L.~M., and Vreeken, J.
\newblock Identifying {{Consistent Statements}} about {{Numerical Data}} with
  {{Dispersion-Corrected Subgroup Discovery}}.
\newblock \emph{Data Mining and Knowledge Discovery}, pp.\  1391--1418,
  September 2017.

\bibitem[Boll \& Lagemann(2019)Boll and Lagemann]{boll:19:paygap}
Boll, C. and Lagemann, A.
\newblock The gender pay gap in eu countries—new evidence based on eu-ses
  2014 data.
\newblock \emph{Intereconomics}, 54:\penalty0 101--105, 2019.

\bibitem[Dierckx et~al.(2023)Dierckx, Veroneze, and Nijssen]{dierckx2023rl}
Dierckx, L., Veroneze, R., and Nijssen, S.
\newblock Rl-net: Interpretable rule learning with neural networks.
\newblock In \emph{Pacific-Asia Conference on Knowledge Discovery and Data
  Mining}, pp.\  95--107. Springer, 2023.

\bibitem[Dinh et~al.(2017)Dinh, Sohl-Dickstein, and Bengio]{dinh:17:density}
Dinh, L., Sohl-Dickstein, J., and Bengio, S.
\newblock Density estimation using real nvp.
\newblock In \emph{International Conference on Learning Representations}, 2017.

\bibitem[Duivesteijn et~al.(2016)Duivesteijn, Feelders, and
  Knobbe]{duivesteijn_exceptional_2016}
Duivesteijn, W., Feelders, A.~J., and Knobbe, A.
\newblock Exceptional {{Model Mining}}: {{Supervised}} descriptive local
  pattern mining with complex target concepts.
\newblock \emph{Data Mining and Knowledge Discovery}, pp.\  47--98, January
  2016.

\bibitem[Durkan et~al.(2019)Durkan, Bekasov, Murray, and
  Papamakarios]{durkan:19:splineflows}
Durkan, C., Bekasov, A., Murray, I., and Papamakarios, G.
\newblock Neural spline flows.
\newblock \emph{Advances in Neural Information Processing Systems}, 32, 2019.

\bibitem[Fischer \& Vreeken(2021)Fischer and
  Vreeken]{fischer2021differentiable}
Fischer, J. and Vreeken, J.
\newblock Differentiable pattern set mining.
\newblock In \emph{Proceedings of the 27th ACM SIGKDD Conference on Knowledge
  Discovery \& Data Mining}, pp.\  383--392, 2021.

\bibitem[Friedman \& Fisher(1999)Friedman and Fisher]{friedman:1999:bump}
Friedman, J.~H. and Fisher, N.~I.
\newblock Bump hunting in high-dimensional data.
\newblock \emph{Statistics and computing}, 9\penalty0 (2):\penalty0 123--143,
  1999.

\bibitem[Giljohann et~al.(2020)Giljohann, Seferos, Daniel, Massich, Patel, and
  Mirkin]{giljohann2020gold}
Giljohann, D.~A., Seferos, D.~S., Daniel, W.~L., Massich, M.~D., Patel, P.~C.,
  and Mirkin, C.~A.
\newblock Gold nanoparticles for biology and medicine.
\newblock \emph{Spherical Nucleic Acids}, pp.\  55--90, 2020.

\bibitem[Goldsmith et~al.(2017)Goldsmith, Boley, Vreeken, Scheffler, and
  Ghiringhelli]{goldsmith2017uncovering}
Goldsmith, B.~R., Boley, M., Vreeken, J., Scheffler, M., and Ghiringhelli,
  L.~M.
\newblock Uncovering structure-property relationships of materials by subgroup
  discovery.
\newblock \emph{New Journal of Physics}, 19\penalty0 (1):\penalty0 013031,
  2017.

\bibitem[Grosskreutz \& R{\"u}ping(2009)Grosskreutz and
  R{\"u}ping]{grosskreutz_subgroup_2009}
Grosskreutz, H. and R{\"u}ping, S.
\newblock On subgroup discovery in numerical domains.
\newblock \emph{Data Mining and Knowledge Discovery}, pp.\  210--226, October
  2009.

\bibitem[Helal(2016)]{helal2016subgroup}
Helal, S.
\newblock Subgroup discovery algorithms: a survey and empirical evaluation.
\newblock \emph{Journal of computer science and technology}, 31:\penalty0
  561--576, 2016.

\bibitem[Hochreiter(1998)]{hochreiter:98:vanishing}
Hochreiter, S.
\newblock The vanishing gradient problem during learning recurrent neural nets
  and problem solutions.
\newblock \emph{International Journal of Uncertainty, Fuzziness and
  Knowledge-Based Systems}, 6\penalty0 (02):\penalty0 107--116, 1998.

\bibitem[Izzo et~al.(2023)Izzo, Liu, and Zou]{izzo2023data}
Izzo, Z., Liu, R., and Zou, J.
\newblock Data-driven subgroup identification for linear regression.
\newblock In \emph{Proceedings of the 40th International Conference on Machine
  Learning (ICML)}, 2023.

\bibitem[Kalofolias et~al.(2019)Kalofolias, Boley, and
  Vreeken]{kalofolias_discovering_2019}
Kalofolias, J., Boley, M., and Vreeken, J.
\newblock Discovering robustly connected subgraphs with simple descriptions.
\newblock In \emph{2019 {{IEEE International Conference}} on {{Data Mining}}
  ({{ICDM}})}, pp.\  1150--1155, November 2019.

\bibitem[Kingma \& Ba(2015)Kingma and Ba]{kingma:15:adam}
Kingma, D.~P. and Ba, J.
\newblock Adam: {A} method for stochastic optimization.
\newblock In Bengio, Y. and LeCun, Y. (eds.), \emph{3rd International
  Conference on Learning Representations, {ICLR} 2015, San Diego, CA, USA, May
  7-9, 2015, Conference Track Proceedings}, 2015.
\newblock URL \url{http://arxiv.org/abs/1412.6980}.

\bibitem[Kl{\"o}sgen(1996)]{klosgen:96:explora}
Kl{\"o}sgen, W.
\newblock Explora: A multipattern and multistrategy discovery assistant.
\newblock In \emph{Advances in knowledge discovery and data mining}, pp.\
  249--271. 1996.

\bibitem[Lavra{\v c} et~al.(2004)Lavra{\v c}, Kav{\v s}ek, Flach, and
  Todorovski]{lavrac_subgroup_2004}
Lavra{\v c}, N., Kav{\v s}ek, B., Flach, P., and Todorovski, L.
\newblock Subgroup discovery with {{CN2-SD}}.
\newblock \emph{Journal of Machine Learning Research}, pp.\  153--188, 2004.

\bibitem[Lemmerich \& Becker(2018)Lemmerich and
  Becker]{lemmerich:2018:pysubgroup}
Lemmerich, F. and Becker, M.
\newblock pysubgroup: Easy-to-use subgroup discovery in python.
\newblock In \emph{Joint European Conference on Machine Learning and Knowledge
  Discovery in Databases}, pp.\  658--662, 2018.

\bibitem[Ortiz \& Cummins(2011)Ortiz and Cummins]{ortiz:11:inequality}
Ortiz, I. and Cummins, M.
\newblock Global inequality: Beyond the bottom billion--a rapid review of
  income distribution in 141 countries.
\newblock \emph{UNICEF, Division of Policy and Strategy Working papers},
  \penalty0 (1102), 2011.

\bibitem[Ouyang et~al.(2018)Ouyang, Curtarolo, Ahmetcik, Scheffler, and
  Ghiringhelli]{ouyang2018sisso}
Ouyang, R., Curtarolo, S., Ahmetcik, E., Scheffler, M., and Ghiringhelli, L.~M.
\newblock Sisso: A compressed-sensing method for identifying the best
  low-dimensional descriptor in an immensity of offered candidates.
\newblock \emph{Physical Review Materials}, 2\penalty0 (8):\penalty0 083802,
  2018.

\bibitem[Papamakarios et~al.(2021)Papamakarios, Nalisnick, Rezende, Mohamed,
  and Lakshminarayanan]{papamakarios:21:normalizing}
Papamakarios, G., Nalisnick, E., Rezende, D.~J., Mohamed, S., and
  Lakshminarayanan, B.
\newblock Normalizing flows for probabilistic modeling and inference.
\newblock \emph{The Journal of Machine Learning Research}, 22\penalty0
  (1):\penalty0 2617--2680, 2021.

\bibitem[Proen{\c{c}}a et~al.(2022)Proen{\c{c}}a, Gr{\"u}nwald, B{\"a}ck, and
  van Leeuwen]{proencca:2022:robust}
Proen{\c{c}}a, H.~M., Gr{\"u}nwald, P., B{\"a}ck, T., and van Leeuwen, M.
\newblock Robust subgroup discovery: Discovering subgroup lists using mdl.
\newblock \emph{Data Mining and Knowledge Discovery}, 36\penalty0 (5):\penalty0
  1885--1970, 2022.

\bibitem[Qiao et~al.(2021)Qiao, Wang, and Lin]{qiao2021learning}
Qiao, L., Wang, W., and Lin, B.
\newblock Learning accurate and interpretable decision rule sets from neural
  networks.
\newblock In \emph{Proceedings of the AAAI Conference on Artificial
  Intelligence}, volume~35, pp.\  4303--4311, 2021.

\bibitem[Rezende \& Mohamed(2015{\natexlab{a}})Rezende and
  Mohamed]{rezende:15:flows}
Rezende, D. and Mohamed, S.
\newblock Variational inference with normalizing flows.
\newblock In \emph{International Conference on Machine Learning}, pp.\
  1530--1538. PMLR, 2015{\natexlab{a}}.

\bibitem[Rezende \& Mohamed(2015{\natexlab{b}})Rezende and
  Mohamed]{rezende:15:variational}
Rezende, D. and Mohamed, S.
\newblock Variational inference with normalizing flows.
\newblock In \emph{International conference on machine learning}, pp.\
  1530--1538. PMLR, 2015{\natexlab{b}}.

\bibitem[Song et~al.(2016)Song, Kull, Flach, and
  Kalogridis]{song_subgroup_2016}
Song, H., Kull, M., Flach, P., and Kalogridis, G.
\newblock Subgroup {{Discovery}} with {{Proper Scoring Rules}}.
\newblock In \emph{Machine {{Learning}} and {{Knowledge Discovery}} in
  {{Databases}}}, pp.\  492--510. {Springer, Cham}, September 2016.
\newblock ISBN 978-3-319-46226-4.

\bibitem[Sutton et~al.(2020)Sutton, Boley, Ghiringhelli, Rupp, Vreeken, and
  Scheffler]{sutton:20:subgroup-material}
Sutton, C., Boley, M., Ghiringhelli, L.~M., Rupp, M., Vreeken, J., and
  Scheffler, M.
\newblock Identifying domains of applicability of machine learning models for
  materials science.
\newblock \emph{Nature communications}, 11\penalty0 (1):\penalty0 4428, 2020.

\bibitem[van Leeuwen \& Knobbe(2011)van Leeuwen and
  Knobbe]{leeuwen_non-redundant_2011}
van Leeuwen, M. and Knobbe, A.
\newblock Non-redundant subgroup discovery in large and complex data.
\newblock In \emph{Machine {{Learning}} and {{Knowledge Discovery}} in
  {{Databases}}}, pp.\  459--474. {Springer}, 2011.
\newblock ISBN 978-3-642-23808-6.

\bibitem[Walter et~al.(2024)Walter, Fischer, and Vreeken]{walter2024finding}
Walter, N.~P., Fischer, J., and Vreeken, J.
\newblock Finding interpretable class-specific patterns through efficient
  neural search.
\newblock In \emph{Proceedings of the AAAI Conference on Artificial
  Intelligence}, 2024.

\bibitem[Wang et~al.(2020)Wang, Zhang, Liu, and Wang]{wang_transparent_2020}
Wang, Z., Zhang, W., Liu, N., and Wang, J.
\newblock Transparent {{Classification}} with {{Multilayer Logical
  Perceptrons}} and {{Random Binarization}}.
\newblock \emph{Proceedings of the AAAI Conference on Artificial Intelligence},
  pp.\  6331--6339, April 2020.

\bibitem[Wang et~al.(2021)Wang, Zhang, Liu, and Wang]{wang2021scalable}
Wang, Z., Zhang, W., Liu, N., and Wang, J.
\newblock Scalable rule-based representation learning for interpretable
  classification.
\newblock \emph{Advances in Neural Information Processing Systems},
  34:\penalty0 30479--30491, 2021.

\bibitem[Yang et~al.(2018)Yang, Morillo, and Hospedales]{yang:18:deeptree}
Yang, Y., Morillo, I.~G., and Hospedales, T.~M.
\newblock Deep neural decision trees.
\newblock \emph{arXiv preprint arXiv:1806.06988}, 2018.

\end{thebibliography}
\newpage
\appendix
\onecolumn
\section{Proof of Asymptotic Correctness of Soft-Binning}
\label{ap:binproof}
\begin{proof}
    Consider a real value $x_i \in \R$ and $M$ sorted bin thresholds $\beta_{i,j} \in \R$, i.e.~$\beta_{i,j} < \beta_{i,j+1}$.
    From the thresholds $\beta_{i,j}$, we construct the bias vector $b_i\in \R^{M+1}$ defined as 
    $$b_i = (0,\sum_{j=1}^1 -\beta_{i,j}, \dots, \sum_{j=1}^M -\beta_{i,j})^T\;.$$
    Additionally, we define a weight vector $w \in \mathbb{R}^{M+1}$ with $w_j = j$, so that $$w = (1,2,\dots,M+1)^T\;.$$

    The soft-binning result $z \in [0,1]^{M+1}$ is defined as 
    \[
        z = \text{softmax}\left((w x_i + b_i)/\lpartemp\right)    \;.
    \]
    Now, let $x_i$ be in the $l$-th bin, i.e.~$\beta_{i,l-1} < x_i < \beta_{i,l}$, then we now firstly prove that 
    $\forall j \neq l: z_l > z_j$.
    We do this by showing that the $l$-th logit $\bar{z}_l = w_l x_i + b_{i,l}$ is the largest and hence also has the 
    highest softmax activation.

    Firstly, note that the bin thresholds are sorted in order, so that for $j<l$ it also holds that $\beta_{i,j} < \beta_{i,l}$.
    $\bar{z}_l$ is defined as
    \[
        \bar{z}_l = w_l x_i + b_{i,l} = w_l x_i - \sum_{k=1}^{l-1} \beta_{i,k}\;.
    \] 
    We can simply transform $\bar{z}_l$ into $\bar{z}_{l-1}$ by subtracting $x_i - \beta_{i,l-1}$, so that
    \[
        \bar{z}_l - x_i + \beta_{i,l}  = w_{l-1} x_i - \sum_{k=1}^{l-2} \beta_{i,k} = \bar{z}_{l-1}\;.
    \]
    Now, as $x_i$ is in the $l$-th bin, we know that $\beta_{i,l-1}<x_i$ and hence $x_i - \beta_{i,l}<0$.
    For all other $j < l$ $\beta_{i,j}<x_i$ holds, and hence also $\bar{z}_l > \bar{z}_j$.
    
    Now consider the case where $j > l$.
    Here, it holds that 
    \[
        \bar{z}_l + x_i - \beta_{i,l+1}  = w_{l+1} x_i - \sum_{k=1}^{l+1} \beta_{i,k} = \bar{z}_{l+1}\;.
    \]
    In general, we may transform $\bar{z}_l$ into $\bar{z}_j$ by repeatedly adding $x_i - \beta_{i,k}$ for $k \in [l+1,\dots,j]$.
    For all thresholds $x_i < \beta_{i,k}$ holds. Thus, each time we add a strictly negative number 
    to the logit $\bar{z}_l$, which proves that also here $\forall j > l: \bar{z}_l > \bar{z}_j$.
    Thus, it holds that $\forall j \neq l: \bar{z}_l > \bar{z}_j$

    Lastly, it remains to prove that with temperature $\lpartemp \to 0$, $z$ is a one-hot bin encoding,
    i.e.~$z_l=1$ and $\forall j \neq l: z_j = 0$.
    The soft-binning of $z_l$ is defined as 
    \[
        \lim_{\lpartemp\to 0} z_l = \lim_{\lpartemp\to 0} \frac{\exp(\bar{z}_l/\lpartemp)}{\sum_{j=1}^{M+1}\exp(\bar{z}_j/\lpartemp)}
        = \lim_{\lpartemp\to 0} \frac{1}{\sum_{j=1}^{M+1}\exp\left((\bar{z}_j-\bar{z}_l)/\lpartemp\right)}\;.
    \]
    For $j = l$, the sum term evaluates to $\exp(\bar{z}_l-\bar{z}_l)/\lpartemp=\exp(0)=1$.
    For $j \neq l$, it holds that $\bar{z}_l>\bar{z}_j$ as show previously, and hence in the limit
    \[
        \lim_{\lpartemp\to 0} \exp(\bar{z}_j-\bar{z}_j)/\lpartemp=\exp(-\infty)=0\;.
    \]
    Thus, in the limit $\lpartemp \to 0$, the denominator sums up to $1$ and hence $z_l = 1$, and as the softmax is 
    positive and sums up to zero, it follows that $\forall j \neq l: z_j = 0$.
\end{proof}

\section{Proof of Theorem \ref{thm:approx}}\label{sec:approx}
\theoremapx
\begin{proof}
  We first recall that, under our model, $\fpdf{\lrsel=1|\lrfeat}(\lfeat)=s_{t\rightarrow0}(\lfeat;\fvec\llb,\fvec\lub)$ for some $\fvec\llb, \fvec\lub\in\lreals^\lnsamp$, and is therefore a smooth function of $\lfeat$.
  Intuitively, there are two regions of interest within $\lfeatspacem$: one within which it transitions from the value of almost $1$ to that of almost $0$, which is the region $\lfeatspacem\setminus\lfeatspacems$, and a saturation region, where $\fpdf{\lrsel=1|\lfeat}\to0$ super-exponentially, which is the region $\lfeatspacems$. The particular thresholds that define these regions are not important, and any reasonable scheme leads to vanishing bounds $\epsilon_1$, $\epsilon_2$.

More formally, using the partitioning of $\lreals^\lnsamp$, we can split the integral of Eq.~\eqref{eq:targsg-pdf} into
\begin{align}
    \lpdftargsg(\ltarg) =  \int_{\lfeat\in\lfeatspacep}\fpdf{\lrtarg|\lrsel=1,\lrfeat}(\ltarg,\lfeat)
  \frac{\fpdf{\lrsel=1|\lrfeat}(\lfeat)\fpdf\lrfeat(\lfeat)}{\fprb{\lrsel=1}}dx + \int_{\lfeat\in\lfeatspacem}\fpdf{\lrtarg|\lrsel=1,\lrfeat}(\ltarg,\lfeat)
  \frac{\fpdf{\lrsel=1|\lrfeat}(\lfeat)\fpdf\lrfeat(\lfeat)}{\fprb{\lrsel=1}}dx \;,
  \label{eq:targsg-splitted}
\end{align}
with the goal to upper bound (and hence ignore) the second term, which we consider as an error.
This second term can be now bounded as
\begin{multline}
\int_{\lfeat\in\lfeatspacem}\fpdf{\lrtarg|\lrsel=1,\lrfeat}(\ltarg,\lfeat)
  \frac{\fpdf{\lrsel=1|\lrfeat}(\lfeat)\fpdf\lrfeat(\lfeat)}{\fprb{\lrsel=1}}d\lfeat \leq\\ \frac1{\fprb{\lrsel=1}}\int_{\lfeat\in\lfeatspacem}C_\lrtarg
  \fpdf{\lrsel=1|\lrfeat}(\lfeat)\fpdf\lrfeat(\lfeat)d\lfeat \leq\\
  \frac{C_\lrtarg}{\fprb{\lrsel=1}}\left[
  \int_{\lfeat\in\lfeatspacem\setminus\lfeatspacems}\underbrace{\fpdf{\lrsel=1|\lrfeat}(\lfeat)}_{\leq 1}\fpdf\lrfeat(\lfeat)d\lfeat  + 
  \int_{\lfeat\in\lfeatspacems}\fpdf{\lrsel=1|\lrfeat}(\lfeat)
  \underbrace{\fpdf\lrfeat(\lfeat)}_{\leq C_\lrfeat}d\lfeat
\right]\leq\\
  \frac{C_\lrtarg}{\fprb{\lrsel=1}}\left[
  \int_{\lfeat\in\lfeatspacem\setminus\lfeatspacems}\fpdf\lrfeat(\lfeat)d\lfeat  + 
  C_\lrfeat\int_{\lfeat\in\lfeatspacems}\fpdf{\lrsel=1|\lrfeat}(\lfeat)d\lfeat
\right]\leq\\
\frac{C_\lrtarg\left(\epsilon_2 + C_\lrfeat\epsilon_1\right)}{\fprb{\lrsel=1}}\;,
\end{multline}
where $\fpdf{\lrsel=1|\lrfeat}\leq1$ since $\lrsel$ is a discrete random variable.

We argue about the second part by claiming that both bounds $\epsilon_1$ and $\epsilon_2$ vanish during learning. Indeed, the form $\lprule(\lfeat)\to\lpdfsel(\lfeat)$ satisfies the assumption of Eq.~\eqref{eq:ass-memb} for a steep enough temperature parameter $\lpartemp$, while it is also learning the correct domain $\lfeatspacem$, so that indeed the assumption of Eq.~\eqref{eq:ass-align} is satisfied, both with inexorably diminishing bounds $\epsilon_1$ and $\epsilon_2$, respectively.
\end{proof}

\newpage
\section{Algorithm and Hyperparameters}
\label{ap:algo}
In this section, we provide pseudocode for \ourmethod.
\begin{algorithm2e}[!h]
    \caption{fit\_flow($\{x^{(1)},\dots,x^{(n)}\}$,$\{y^{(1)},\dots,y^{(n)}\}$,$s$,$\fpdf{}$,$\bar{p}$)}
    $\log \mathcal{L} \gets \frac{1}{n} \sum_{i=1}^n \log[\fpdf{}(y^{(i)})\cdot s(x^{(i)})+\bar{p}(y^{(i)})\cdot (1-s(x^{(i)}))]$\;
    loss $\gets -\log \mathcal{L}$\;
    loss.$\text{backwards}()$\;
    Update $\fpdf{}$\;
\end{algorithm2e}

\begin{algorithm2e}[!h]
    \caption{SYFLOW($X$, $Y$, $ \text{epochs}_{\text{Flow}_{\text{Y}}}$, $\text{epochs}_{\text{Flow}_{\text{Y}_\text{s}}}$, $\text{lr}_{\text{Flow}}$ ,$\text{lr}_{\text{s}}$,$\text{priors}$, $\gamma$, $\lambda,\lpartemp$)}
    $\fpdf Y \gets \text{Neural Spline Flow}$\;
    \For{$i \gets 1$ \KwTo $ \text{epochs}_{\text{Flow}_{\text{Y}}}$}{
        $\log \mathcal{L} \gets \frac{1}{n} \sum_{i=1}^n \log[\lpdftarg(y^{(i)})]$\;
        loss $ \gets -\log \mathcal{L}$\;
        loss.$\text{backwards}()$\;
        Update $\lpdftarg$\;
    }
    $\alpha_i \gets \min X_i$\;
    $\beta_i \gets \max X_i$\;
    $a_i \gets 1$\;
    Rule$(x)$ $\gets s(x;\alpha,\beta,a,\lpartemp)$\;
    $\lpdftargsg \gets \text{Neural Spline Flow}$\;
    $p_{Y|S=0} \gets \text{Neural Spline Flow}$\;
    \For{$i \gets 1$ \KwTo $\text{epochs}_{\text{Flow}_{\text{Y}}}$ }{
        Compute subgroup membership probabilities $s(x^{(i)})$\;
        KL $ \gets \frac{1}{n} \sum_{i=1}^n s(x^{(i)}) \cdot (\log[\lpdftargsg(y^{(i)})] - \log[\lpdftarg(y^{(i)})])$\;
        $n_s \gets \frac{1}{n} \sum_{i=1}^n s(x^{(i)})$\;
        Weighted-$KL \gets KL \cdot \bar{s}^{\gamma}$\;
        Regularizer $ \gets 0$\;
        \For{$p_{SG_k}$ \textbf{in} priors}{
            Regularizer $\gets$ Regularizer + $\sum_{i=1}^n s(x^{(i)}) \cdot (\log[\lpdftargsg(y^{(i)})] - \log[p_{Y|SG_k=1}(y^{(i)})])$\;
        }
        Regularizer$ \gets \frac{\lambda}{|priors|} \cdot $Regularizer  \;
        loss $ \gets-$ Weighted-KL $- $Regularizer\;
        loss.$\text{backwards}()$\;
        Update rule $s$ to minimize objective/maximize weighted, regularized KL\;
        fit\_flow($\{x^{(1)},\dots,x^{(n)}\}$,$\{y^{(1)},\dots,y^{(n)}\}$,$s$,$\lpdftargsg$,$p_{Y|S=0}$)\;
        \If{$(\lpartemp = \text{epochs}_{\text{Flow}_{\text{Y}_\text{s}}}/2) \vee (\lpartemp = \text{epochs}_{\text{Flow}_{\text{Y}_\text{s}}}\cdot 3/4$)}{
            $\lpartemp \gets \lpartemp/2$\;
        }
        
    }
    \Return Rule, $\lpdftargsg$\;
\end{algorithm2e}
\newpage
\section{Hyperparameters for experimental evaluation}
\label{ap:hyperparameters}
For all methods we optimize their respective hyperparameters such to maximize the
measures used to evaluate the experiments. Since, \bh has no hyperparameters no 
fine-tuning is required.
\subsection{Synthetic experiments}
We used one hyperparameter setting for all synthetic experiments
For \ourmethod the hyperparameter setting is:
$t=0.2$, $\gamma=0.5$, $\lambda=0.5$, $\text{lr}_{\text{Flow}}=\num{5e-2}$, $\text{lr}_{\text{s}}=\num{2e-2}$, $\text{epochs}_{\text{Flow}_{\text{Y}}}=2000$ and $\text{epochs}_{\text{Flow}_{\text{Y}_\text{s}}}=1500$.
For \ssd, \sdkl and \rsd, we used 20 cutpoints and  a beamwidth of 100, while $\gamma$ is set to 1.0.
Although $\gamma$ has for all approaches the same functionality i.e. control the size of the subgroup,
the absolute value of a $\gamma$ has drastically different meanings.
As \ourmethod outputs a soft  asignment, we require a smaller alpha to achieve the same effect.
For example, if \ourmethod assigns $0.9$, then contribution to the size correction for $\gamma=0.5$ is $\sqrt{0.9}\approx 0.95$,
thus we require a smaller $\gamma$.

\subsection{Real world data}
For the experiments on Kaggle and UCI data (i.e. Section \ref{sec:realworld}), we used for 
\ourmethod: $t=0.2$, $\gamma=0.3$, $\lambda=2.0$, $\text{lr}_{\text{Flow}}=\num{5e-2}$, $\text{lr}_{\text{s}}=\num{2e-2}$, $\text{epochs}_{\text{Flow}_{\text{Y}}}=1000$ and $\text{epochs}_{\text{Flow}_{\text{Y}_\text{s}}}=1000$.
For \ssd, \sdkl and \rsd, we used 20 cutpoints, a beamwidth of 100, and while $\gamma=1.0$.
For the experiments on  gold cluster data present in Figure \ref{fig:homo-lumo}, we used
$\text{epochs}_{\text{Flow}_{\text{Y}}}=7000$, $\text{epochs}_{\text{Flow}_{\text{Y}_\text{s}}}=3000$, $\lambda=10.0$ and $\gamma=0.2$
For the case study, we used $\text{epochs}_{\text{Flow}_{\text{Y}}}=3000$, $\text{epochs}_{\text{Flow}_{\text{Y}_\text{s}}}=2000$, $\lambda=5$ and $\gamma=0.2$.

\section{Evaluation metric}
\label{app-sec:metrics}
To objectively evaluate the discovered subgroups on real-world data, we use Bhattacharyya coefficient (BC),
KL-divergence (KL) and absolute mean distance (AMD) between  the 
distribution of the subgroup $P_{Y|S=1}$ and overall distribution of $P_Y$. KL and AMD are size corrected, since both metrics are heavily
influenced by the size of the subgroup. We use histogramms to estimate the  probabilitiy distributions.
The edges of the bins are computed using the Freedman Diaconis Estimator, which is robust to outliers and 
less sensitive towards distribution shapes. For the subgroup distribution $P_{Y|S=1}$ and overall distribution
$P_Y$, the metrics are formally defined as
\begin{align*}
    \BC (P_{Y|S=1}, P_{Y}) &= \sum_{y \in \ltargspace} \sqrt{p_{Y|S=1}(y)p_{Y}(y)} \\
    \DKL (P_{Y|S=1}, P_{Y}) &= \sum_{y \in \ltargspace} p_{Y|S=1}(y) \log (\frac{p_{Y|S=1}(y)}{p_{Y}(y)}) \\
    \AMD (\ltargspace_s, \ltargspace) &= \left| \left(\frac{1}{|\ltargspace_s|}\sum_{y \in \ltargspace_s} y\right) - \left(\frac{1}{|\ltargspace|}\sum_{y \in \ltargspace} y \right) \right| \; .
\end{align*}
In the last defnition, with slight abuse of notation, we used $\ltargspace_s$ to denote all points in the subgroup.
For the Table \ref{tab:real-world}, we size corrected KL and AMD using $\gamma=1$. Note, this is 
exactly the metric that \sd and \sdkl optimize.

\section{Rule Complexity Experiment}
\label{ap:rulecomplexity}
We study how \ourmethod learns subgroups of increasing complexity.
This is achieved by increasing the number of predicates in a generated rule up to ten. We keep the target distribution fixed to a Normal distribution.

\begin{figure}
    \centering
    \begin{tikzpicture}
        \usetikzlibrary{calc}
        \begin{axis}[ 
            pretty line,
            legend style={xshift=-1cm,yshift=1cm,font=\scriptsize},
            cycle list name = prcl-line,
            width=0.5\linewidth,
            height=3.5cm,
            ylabel={F1-score},
            xlabel={Number of predicates in rule},
            pretty labelshift,
            ymax=1.0,
            tick label style={/pgf/number format/fixed},
            legend columns = 7,
            legend entries = {\ourmethod, \sd,\sdkl,\rsd,\bh},
        ]
        \foreach \x in {syflow, sd-mean, sd-kl, rsd, bh}{
            \addplot+[error bars/.cd, 
            y fixed,
            y dir=both, 
            y explicit] table[x={n-rules},y={F1},col sep=comma,y error plus expr=\thisrow{F1-STD}, y error minus expr=\thisrow{F1-STD}]{expres/rule_complexity/\x.csv};
        }      
        \end{axis}
      \end{tikzpicture}
      \caption{F1 under increasingly complex rules, higher is better }
      \label{fig:complexity}
\end{figure}
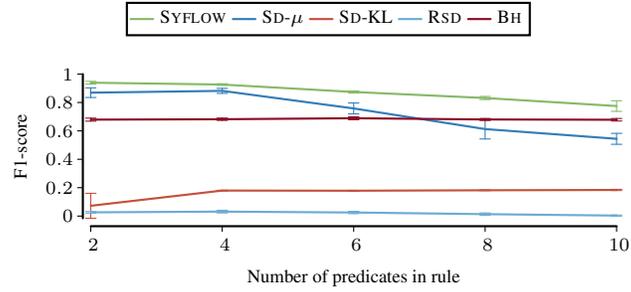

We present our results in Figure \ref{fig:complexity};
Here, as the complexity of the rule increases, the accuracy of all methods generally decreases as the task becomes progressively harder.
Amongst all methods, \ourmethod recovers the planted subgroup with the highest accuracy.
In particular, \ourmethod improves the most over its competitors on the more complex subgroups.

\section{Target distributions for synthetic experiment}
\label{app:target}
\begin{figure*}[!ht]
    \centering
    \pgfplotstableread[col sep=comma]{expres/true_targets.csv}\datatable 
      \begin{subfigure}{0.25\textwidth}
        \begin{tikzpicture}
          \begin{axis}[
            pretty line,
            xlabel={$x$},
            ylabel={$p_{Y|S=1}(y)$},
            width=\textwidth
            ]
            \addplot[blue, thick] table[x=x_normal, y=y_normal] {\datatable};
          \end{axis}
        \end{tikzpicture}
        \caption{$\mathcal{N}(1.5,0.5)$}
      \end{subfigure}%
      \begin{subfigure}{0.25\textwidth}
        \begin{tikzpicture}
          \begin{axis}[
            pretty line,
            xlabel={$x$},
            ylabel={$p_{Y|S=1}(y)$},
            width=\textwidth
            ]
            \addplot[blue, thick] table[x=x_beta, y=y_beta] {\datatable};
          \end{axis}
        \end{tikzpicture}
        \caption{$\mathcal{B}(0.2,0.2)$}
      \end{subfigure}%
      \begin{subfigure}{0.25\textwidth}
        \begin{tikzpicture}
          \begin{axis}[
            pretty line,
            xlabel={$x$},
            ylabel={$p_{Y|S=1}(y)$},
            width=\textwidth
            ]
            \addplot[blue, thick] table[x=x_cauchy, y=y_cauchy] {\datatable};
          \end{axis}
        \end{tikzpicture}
        \caption{$\mathcal{C}(0,1)$}
      \end{subfigure}%
      \begin{subfigure}{0.25\textwidth}
        \begin{tikzpicture}
          \begin{axis}[
            pretty line,
            xlabel={$x$},
            ylabel={$p_{Y|S=1}(y)$},
            width=\textwidth
            ]
            \addplot[blue, thick] table[x=x_expon, y=y_expon] {\datatable};
          \end{axis}
        \end{tikzpicture}
        \caption{$Exp(0.5)$}
      \end{subfigure}

      \begin{subfigure}{0.25\textwidth}
        \begin{tikzpicture}
          \begin{axis}[
            pretty line,
            xlabel={$x$},
            ylabel={$p_{Y|S=1}(y)$},
            width=\textwidth
            ]
            \addplot[blue, thick] table[x=x_rayleigh, y=y_rayleigh] {\datatable};
          \end{axis}
        \end{tikzpicture}
        \caption{$\mathcal{R}(2)$}
      \end{subfigure}%
      \begin{subfigure}{0.25\textwidth}
        \begin{tikzpicture}
          \begin{axis}[
            pretty line,
            xlabel={$x$},
            ylabel={$p_{Y|S=1}(y)$},
            width=\textwidth
            ]
            \addplot[blue, thick] table[x=x_bi, y=y_bi] {\datatable};
          \end{axis}
        \end{tikzpicture}
        \caption{$\mathcal{N}(-1.5,0.5)+\mathcal{N}(1.5,0.5)$}
      \end{subfigure}%
      \begin{subfigure}{0.25\textwidth}
        \begin{tikzpicture}
          \begin{axis}[
            pretty line,
            xlabel={$x$},
            ylabel={$p_{Y|S=1}(y)$},
            width=\textwidth
            ]
            \addplot[blue, thick] table[x=x_uniform, y=y_uniform] {\datatable};
          \end{axis}
        \end{tikzpicture}
        \caption{$\mathcal{U}(0.5,1.5)$}
      \end{subfigure}%
    \caption{Here, we show the different target distributions for second synthetic experiment in Section \ref*{sec:experiments}.}
    \label{appfig:target}
  \end{figure*}
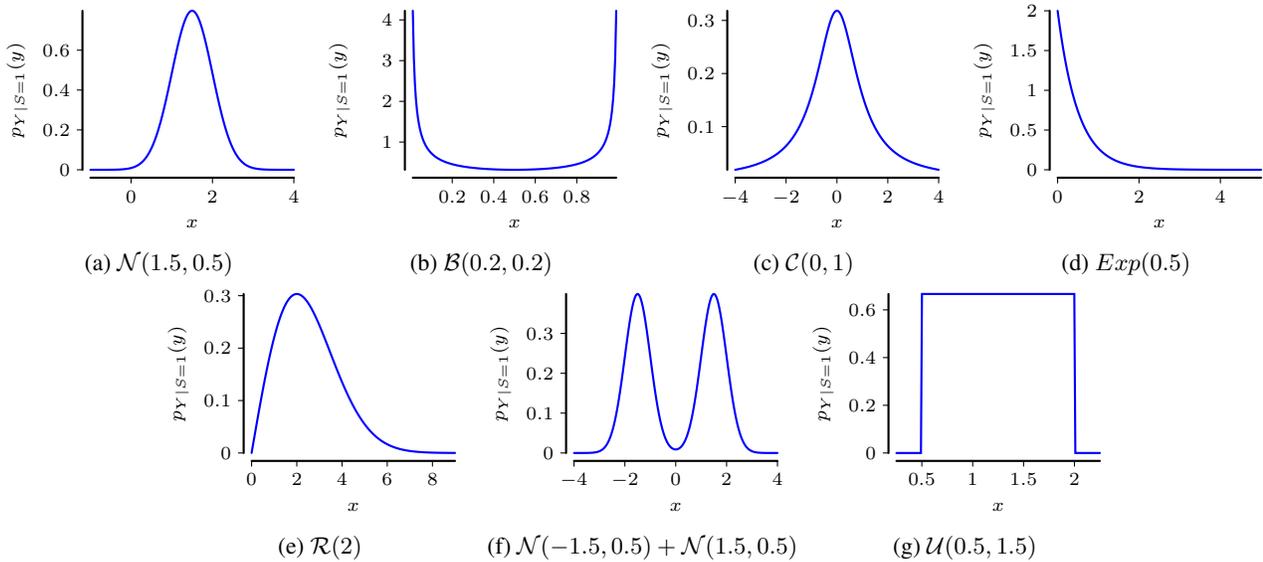

In Figure \ref{appfig:target}, we show the different target distributions for the first synthetic experiment.

\section{Subgroup descriptions}
We show in Table \ref{tab:rules} further examples of subgroups found on the real life datasets. For each method we select the first subgroup 
that the respective method found. Since \bh did not find relevant subgroups for most datasets, we do not consider it in the table.

\begin{table*}
    \centering
    \renewcommand{\arraystretch}{1.5}
    \scriptsize
    \begin{tabular}{llc}
        \toprule
            Dataset & Method  & Rule  \\
            \midrule
            \multirow{4}{*}{\vspace{-0.5cm}Abalone}     &  \ourmethod & 0.20 $<$ whole-weight $<$ 2.83 \& 0.13 $<$ viscera-weight $<$ 0.60 \& 0.26 $<$ shell-weight $<$ 1.00 \& sex-I =0  \\
                                        &  \rsd & \makecell{ shell-weight $\geq$ 0.43 \& 0.46 $\leq$ diameter $<$ 0.49 \& shucked-weight $<$ 0.55 \& whole-weight $\geq$ 1.03}  \\
                                        &  \sdkl & shell-weight $<$ 0.48 \\
                                        &  \ssd & \makecell{height $\geq$ 0.10 \& shucked-weight $\geq$ 0.16 \& shell-weight $\geq$ 0.11 \& shell-weight $\geq$ 0.19} \\
            \midrule
            \multirow{4}{*}{\vspace{-0.5cm}Airquality}     &  \ourmethod & 404.67 $<$ NOx(GT) $<$ 1479.00 \& -44.90 $<$ NO2(GT) $<$ 340.00   \\
                                        &  \rsd & \makecell{ PT08.S4(NO2) $\geq$ 2026.0 \& PT08.S3(NOx) $<$ 373.0 \& month $\geq$ 11.0}  \\
                                        &  \sdkl & \makecell{PT08.S1(CO) $<$ 1285.0 \& NMHC(GT) $<$ 185.0 \& PT08.S2(NMHC) $<$ 1068.0 \& NOx(GT) $<$ 326.0 \\ \& NOx(GT) $\geq$ -200.0 \& PT08.S5(O3) $<$ 1780.0}\\
                                        &  \ssd & \makecell{PT08.S1(CO) $\geq$ 956.0 \& PT08.S2(NMHC) $\geq$ 979.0 \& PT08.S5(O3) $\geq$ 704.0 \& PT08.S5(O3) $\geq$ 917.0} \\
            \midrule
            \multirow{4}{*}{\vspace{-0.5cm}Automobile}     &  \ourmethod & \makecell{If 159.42  $<$  length  $<$  208.10 \& 60.64  $<$  width  $<$  72.00 \&  2684.19  $<$  curb-weight  $<$  4066.00 \& 64.84  $<$  engine-size  $<$  270.46\\ \& 7.00  $<$  compression-ratio  $<$  19.11 \& 16.00  $<$  highway-mpg  $<$  31.72 \\ \&  jaguar=0 \&  mercedes-benz=0 \&  plymouth=0 \&   subaru=0 \& fuel-type-gas=1 \&  engine-location-front=1 \\ \& engine-location-rear=0 \& engine-type-ohcf=0 \& fuel-system-2bbl=0\& fuel-system-mpfi=1 }           \\
                                        &  \rsd & \makecell{ engine-size $\geq$ 201.5}  \\
                                        &  \sdkl & \makecell{highway-mpg $\geq$ 29.0 \&  audi=0 \&  bmw=0 \& wheel-base $<$ 104.90 \& width $<$ 66.90 \\ \& body-style-convertible=0 \& engine-size $<$ 161.0}\\
                                        &  \ssd & \makecell{highway-mpg $<$ 30.0 \&  honda=0 \&  isuzu=0 \&  plymouth=0 \&  subaru=0 \& curb-weight $\geq$ 2385.0 \& fuel-system-mfi=0} \\
            \midrule
            \multirow{4}{*}{\vspace{-0.5cm}Bike}     &  \ourmethod &  holiday=0 \& 1.00  $<$  weathersit  $<$  2.35 \& 0.47  $<$  atemp  $<$  0.78 \& 0.18  $<$  hum  $<$  0.83 \& season-1=0  \\
                                        &  \rsd & \makecell{ temp $<$ 0.2}  \\
                                        &  \sdkl & temp $\geq$ 0.26 \& atemp $\geq$ 0.28 \& hum $<$ 0.87 \& windspeed $<$ 0.34 \& season-1=0 \\
                                        &  \ssd & \makecell{temp $\geq$ 0.40 \& temp $\geq$ 0.43 \& hum $<$ 0.82} \\
            \midrule
            \multirow{4}{*}{\vspace{-0.5cm}California}     &  \ourmethod & 0.50  $<$  MedInc  $<$  2.4  \\
                                        &  \rsd & \makecell{ MedInc $\geq$ 7.4 \& 35.0 $\leq$ HouseAge $<$ 38.0 \& -121.19 $\leq$ Longitude $<$ -118.34 \& AveBedrms $\geq$ 0.96}  \\
                                        &  \sdkl & MedInc $<$ 5.11 \& MedInc $<$ 6.16 \& HouseAge $<$ 52.0 \& AveOccup $\geq$ 2.08 \\
                                        &  \ssd & \makecell{MedInc $\geq$ 3.32 \& AveOccup $<$ 3.89 \& AveOccup $<$ 4.33 \& Latitude $<$ 37.99 \& Longitude $<$ -117.08} \\

            \midrule
            \multirow{4}{*}{\vspace{-0.5cm}Insurance}     &  \ourmethod & 44.00  $<$  age  $<$  64.00 \& smoker=0 \\
                                        &  \rsd & \makecell{ smoker=1 \& bmi $\geq$ 30.0 \& age $\geq$ 59.0}  \\
                                        &  \sdkl & smoker=0 \\
                                        &  \ssd & \makecell{smoker=1} \\
            \midrule
            \multirow{4}{*}{\vspace{-0.5cm}Mpg}     &  \ourmethod & 3.35  $<$  cylinders  $<$  5.25 \\
                                        &  \rsd & \makecell{ weight $\geq$ 3845.0 \& 71.0 $\leq$ model-year $<$ 74.5}  \\
                                        &  \sdkl & displacement $\geq$ 151.0 \& weight $\geq$ 2671.0 \& weight $\geq$ 3085.0 \\
                                        &  \ssd & \makecell{cylinders=4.0 \& weight $<$ 2807.0 \& weight $<$ 4278.0} \\

            \midrule
            \multirow{4}{*}{\vspace{-0.5cm}Student}     &  \ourmethod & \makecell{school=1 \& address=0 \&  failures=0 \& schoolsup=1 \& nursery=0 \& \\ higher=0 \& internet=1\& 1.00  $<$  Dalc  $<$  3.34 \& Medu-1=0\& Fedu-1=0}  \\
                                        &  \rsd & \makecell{ failures =1 \& famsize=1 \& absences $<$ 1 \& famsup=0}  \\
                                        &  \sdkl & failures=0 \\
                                        &  \ssd & \makecell{school=0 \& higher=0 \& absences $<$ 16 \& absences $<$ 18 \& failures=0 \& schoolsup=1} \\

                \midrule
            \multirow{4}{*}{Wages}  &  \ourmethod &  sex=0 \& 10.68  $<$  ed  $<$  18.00 \& 29.86  $<$  age  $<$  95.00 \\
                                        &  \rsd & ed $\geq$ 18.0 \& 37.0 $\leq$ age $<$ 43.0 \& sex=0 \& height $<$ 70.25 \\
                                        &  \sdkl & race-other=0.0 \\
                                        &  \ssd & height $\geq$ 65.05 \& height $\geq$ 65.79 \& sex=0 \& age $\geq$ 30.0 \\
                \midrule
            \multirow{4}{*}{Wine}  &  \ourmethod &  0.23  $<$  volatile acidity  $<$  1.10 \& 0.99  $<$  density  $<$  1.04 \& 8.00  $<$  alcohol  $<$  10.12 \\
                                        &  \rsd & alcohol $\geq$  12.75  \&  29.0 $<$ free sulfur dioxide $<$ 41.0  \&  pH $<$ 3.24  \&  0.26 $\leq$ citric acid<0.34  \&  residual sugar $\geq$  1.4 \\
                                        &  \sdkl & free sulfur dioxide $\geq$ 11.0 \\
                                        &  \ssd & fixed acidity $<$ 8.30 \& alcohol $\geq$ 10.40 \& free sulfur dioxide $\geq$ 11.0 \& total sulfur dioxide $<$ 195.0 \& density $<$ 1.00  \\

        \bottomrule
    \end{tabular}
    \caption{Symbolic subgroup descriptions for real life datasets in Section \ref{sec:realworld}}
    \label{tab:rules}
\end{table*}

\end{document}